\documentclass[runningheads]{llncs}

\usepackage{eccv}

\usepackage{eccvabbrv}

\usepackage{graphicx}
\usepackage{booktabs}
\usepackage{multirow}
\usepackage{array}
\usepackage{overpic}
\usepackage{color}
\usepackage{epstopdf}
\usepackage{wrapfig}
\usepackage[accsupp]{axessibility}  

\usepackage{hyperref}

\usepackage{orcidlink}
\begin{document}

\title{ShoeModel: Learning to Wear on the User-specified Shoes via Diffusion Model} 

\titlerunning{Abbreviated paper title}

\author{Binghui Chen$^{*}$ \and
Wenyu Li\inst{1}$^{*}$\and
Yifeng Geng \and Xuansong Xie \and  Wangmeng Zuo\inst{1}}

\authorrunning{F.~Author et al.}

\institute{$^{*}$Equal contribution\\$^{1}$Harbin Institute of Technology, China\\
\email{chenbinghui@bupt.cn, liwenyu27@gmail.com, cangyu.gyf@alibaba-inc.com, xingtong.xxs@taobao.com, wmzuo@hit.edu.cn}}

\maketitle

\begin{figure}
  \includegraphics[width=1.0\linewidth]{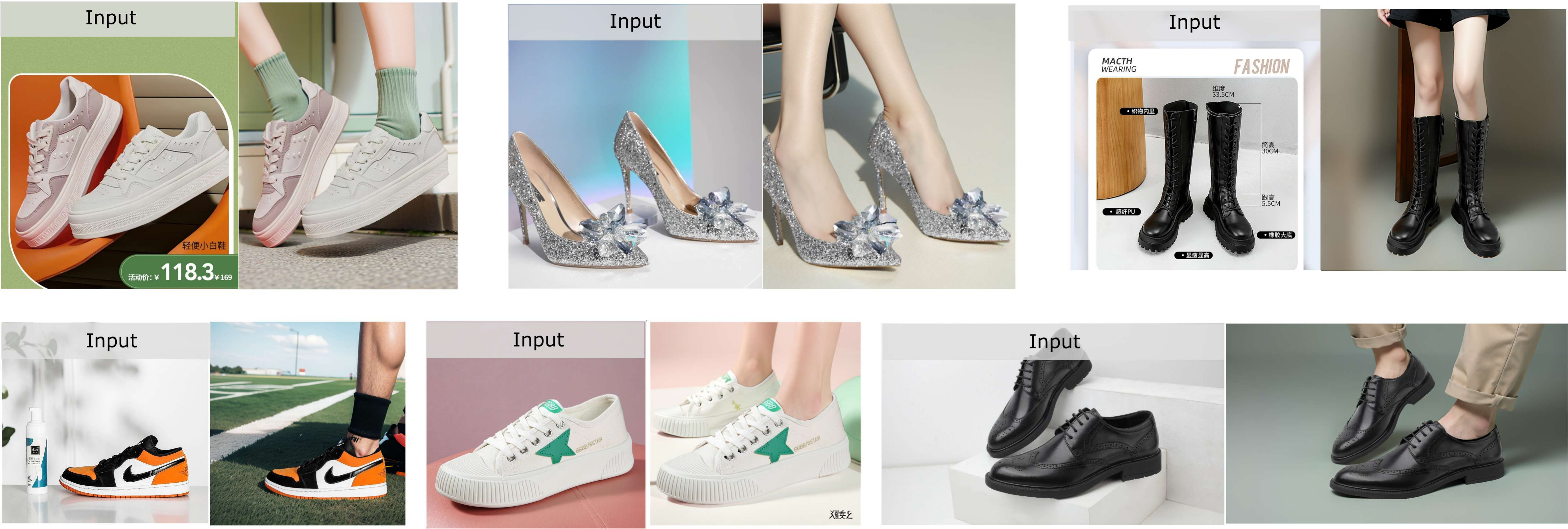}
  \caption{\small{\textbf{Example Results of the ShoeModel.} Each group contains the input user-specified shoes (at left) and the output images interacting with human-legs (at right). It can be observed that (1) the generated images are hyper-realistic and good for product shown, (2) the interaction between human and shoes is reasonable, (3) the identity of the output shoes is exactly the same as the intput ones. \textbf{Best viewed with zoom-in}.}}\label{fig_intro}
\end{figure}

\begin{abstract}
  With the development of the large-scale diffusion model, Artificial Intelligence Generated Content (AIGC) techniques are popular recently. However, how to truly make it serve our daily lives remains an open question. To this end, in this paper, we focus on employing AIGC techniques in one filed of E-commerce marketing, \ie, generating hyper-realistic advertising images for displaying user-specified shoes by human. Specifically, we propose a shoe-wearing system, called \textbf{ShoeModel}, to generate plausible images of human legs interacting with the given shoes. It consists of three modules: (1) shoe wearable-area detection module (WD), (2) leg-pose synthesis module (LpS) and the final (3) shoe-wearing image generation module (SW). Them three are performed in ordered stages. Compared to baselines, our ShoeModel is shown to generalize better to different type of shoes and has ability of keeping the ID-consistency of the given shoes, as well as automatically producing reasonable interactions with human. Extensive experiments show the effectiveness of our proposed shoe-wearing system. Figure \ref{fig_intro} shows the input and output examples of our ShoeModel.
\end{abstract}

\section{Introduction}
\label{sec:intro}
With the help of the pre-trained stable diffusion model \cite{rombach2022high}, many multi-modal conditional image synthesis methods \cite{zhang2023adding, mou2023t2i, ju2023humansd,zhang2023controlvideo,xue2024strictly,chen2024virtualmodel} have ability to generate controllable images. These methods aim to precisely control the generation process. However, these methods cannot preserve the identity of the objects in the conditional images which is crucial for the synthesis of realistic electronic business advertising images. Moreover, the interaction between conditional objects and synthesis human is also not considered in these existing methods.

Thus, in this paper, we consider the shoe-wearing image synthesis scenario,which is commonly encountered in electronic business but has not been thoroughly explored or implemented before. We propose a \emph{ShoeModel} system, which can generate reasonable and satisfied quality advertising images of shoes worn by a human-model, in which shoes are exactly same with the original user-specified ones. Considering two main challenges that have not been explored in the existing controllable image generation works: (1) keeping the identity of the user-specified objects (shoes in this paper) unchanged after the generation; (2) generating plausible interactions between human-models and the conditional objects (shoes in this paper) without extra human-pose inputs, our proposed \emph{ShoeModel} introduces three modules to solve these above problems, including the shoe wearable-area detection module (WD), the leg-pose synthesis module (LpS) and the final shoe-wearing image generation module (SW).

Specifically, in the WD module, we define the visible areas of the shoes as the regions that remain visible whether the shoes are worn or not, while the wearable areas are only visible when the shoes are not worn(referring to Figure \ref{fig:shoe wearable area example}). After training WD to distinguish them, we can get a processed shoe image that retains only the visible areas while removing the background and wearable areas, such that the human-model can interact with this shoe (\ie, wearing this shoe) without any sense of disharmony. Furthermore, recognizing the multitude of potential human poses for a given shoe image, we employ a diffusion model in the LpS module. The meticulously processed shoe image serves as the condition for this module. Consequently, the LpS module is capable of generating diverse and plausible human-leg poses. In the final SW module, we generate desired shoe-wearing images with the constraint of the visible area of the given shoes and the synthesised leg-pose. This generation can both keep the identity of the user-specified shoes unchanged and produce realistic interactions between the human-model and the given shoes.

The contributions of this paper can be summarized as:
\begin{enumerate}
    \item We propose a \emph{ShoeModel} system which is the first focusing on the shoe-wearing image synthesis that prioritize the preservation of the given shoes' identity and the generation of realistic interactions between the shoes and the human- model.
    \item \emph{ShoeModel} consists of the shoe wearable-area detection module (WD), the leg-pose synthesis module (LpS), and the final shoe-wearing image generation module (SW).
    \item We construct a shoe-wearing dataset to support the training of each modules above. It consists of a wearable-area detection sub-dataset and a shoe-leg synthesis sub-dataset.
\end{enumerate}
Extensive experiments have been conducted to demonstrate the effectiveness of the proposed \emph{ShoeModel} system.

\section{Related Works}
\label{sec:related}
\textbf{Text-to-image Diffusion model.}
\label{subsec:T2I DM}
Text-to-image (T2I) generation task aims to generate high-quality synthesis images related to the providing text information. In recent years, with the development of diffusion models \cite{nichol2021glide, rombach2022high, ramesh2022hierarchical}, T2I tasks have made rapid progress. GLIDE \cite{nichol2021glide} compares Classifier-free diffusion guidance metric \cite{ho2021classifier} and CLIP \cite{radford2021learning} guidance metric, and shows that CLIP \cite{radford2021learning} guidance metric can generate images more related to the text guidance. The following approaches, \emph{e.g.} DALL-E2 \cite{ramesh2022hierarchical} and Stable Diffusion \cite{rombach2022high}, achieve impressive improvement through the use of CLIP \cite{radford2021learning} and their proposed frameworks. Nowadays, DALL-E2 \cite{ramesh2022hierarchical} and Stable Diffusion \cite{rombach2022high} have become widely used base model attracting researchers to conduct research on the basis of them. However, these methods are still ineffective in generating real human body images and do not support the interaction with the given objects.

\textbf{Controllable Diffusion model}
\label{subsec:controllable DM}
In order to perform precise control over the generation, researchers focus on adding multi-modal guidance into the T2I generation process. GLIGEN \cite{li2023gligen} introduces keypoint and bounding box information with specific captions into the pre-trained T2I diffusion model \emph{e.g.} SD model \cite{rombach2022high} through their proposed Gated Self-Attention module so that the position and pose of the generated object/human can be controlled. ControlNet \cite{zhang2023adding} and T2I-adapter \cite{mou2023t2i} propose additional trainable modules so that sketch information and human pose information can guide the generation of the pre-trained SD model \cite{rombach2022high} through this module. On the other hand, instead of using additional trainable modules, HumanSD \cite{ju2023humansd} proposes a heatmap-guided denoising loss to control learning on humans. But they still need the pose inputs to generate human bodies. At the same time, they do not pay enough attention to the interaction between human and objects, especially when a specific object is given and a human needs to be generated to interact with it, which is a common scenario in E-commerce marketing. Moreover, the situation of keeping the identity of a specific object unchanged is also not considered by the methods introduced above.

In addition, we noticed that both training-based methods like DreamBooth \cite{ruiz2023dreambooth} and training-free methods like IP-adapter \cite{ye2023ip-adapter}, claim to be capable of keeping the identity of specified objects. However, most of them mainly show the examples of person or animals or cartoon human which have one thing in common, \ie, these are visually insensitive to subtle visual changes. When being applied in e-commerce scenario, their performances are worse since the identity of the given object cannot be exactly guaranteed. In contrast, painting-based methods like Paint-by-example \cite{yang2023paint} and ControlNet-inpaint \cite{zhang2023adding} will always generate something new contents around the given objects and cannot produce reasonable interactions with human. All the above are verified in section \ref{sec_qualitative}. 
\section{Method}
\label{sec:method}

\begin{figure*}[!t]
  \centering
  \includegraphics[width=0.95\linewidth]{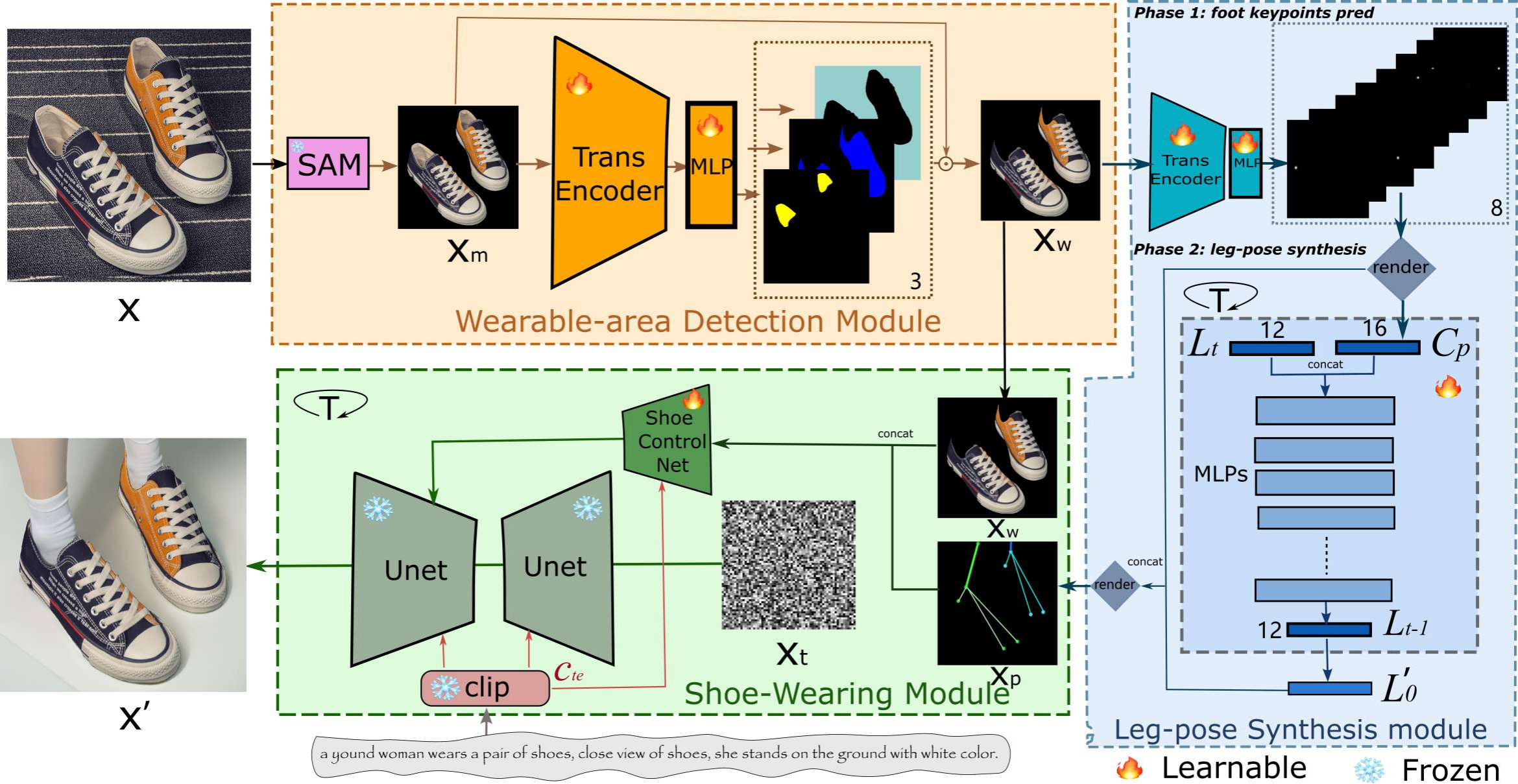}
  \caption{\small{The overall pipeline of the proposed system \emph{ShoeModel}. It consists of three modules: including the shoe Wearable-area Detection module (WD), the Leg-pose Synthesis module (LpS) and the Shoe-wearing module (SW). These three modules are performed in ordered stages and each one can be trained independently. Processed by our system, the given shoes are interacted with human legs reasonably, resulting in a advertising image of shoe display.}}\label{fig:pipline}
\end{figure*}
Given an image of shoe, we aim to synthesize advertising images depicting plausible ways of a human leg interacting with it. And it's worth noting that the identity of shoe in the final generated image should be exactly the same with the input.

To this end, in this section, we introduce the detail of our proposed \emph{ShoeModel} system. The whole pipeline of the proposed fromework is presented in Figure \ref{fig:pipline}. It consists of three modules: the shoe wearable-area detection module (WD, in section\ref{subsec:shoe wearable}), the leg-pose synthesis module (LpS, in section \ref{subsec:leg pose}), and the shoe-wearing image generation module (SW, in section \ref{subsec:leg pose}). And these modules are performed in order. Specifically, the input of the whole pipeline is a user-specified shoe image $X$ and a user-specified text prompt which described the look like of the final generated image. Text prompt will be encoded to embeddings $C_{te}$ by CLIP \cite{radford2021learning}. We first need to obtain the wearable area of the input shoe image, \ie,$X_{w}$. Then, reasonable leg-pose image $X_{p}$ will be synthesized conditioned on the visible area of the input shoe $X_{v}$. Finally, the visible area of shoe $X_{v}$ and the synthesized leg-pose image $X_{p}$ as well as the user-specified prompt condition $C_{te}$ are combined together to guide the generation of shoe shown image by human-model. Each stage can be trained independently.

Our proposed \emph{ShoeModel} takes into consideration the interaction between the input shoe image and human model, the characteristics of human body structure and the identity-consistency between input and output shoe. As as result, it can produce plausible images $X^{'}$ of human legs interacting with the given shoes.


\subsection{Wearable-area Detection Module}
\label{subsec:shoe wearable}
\begin{figure}[!t]
  \centering
  \begin{subfigure}{0.26\linewidth}
    \includegraphics[width=\linewidth]{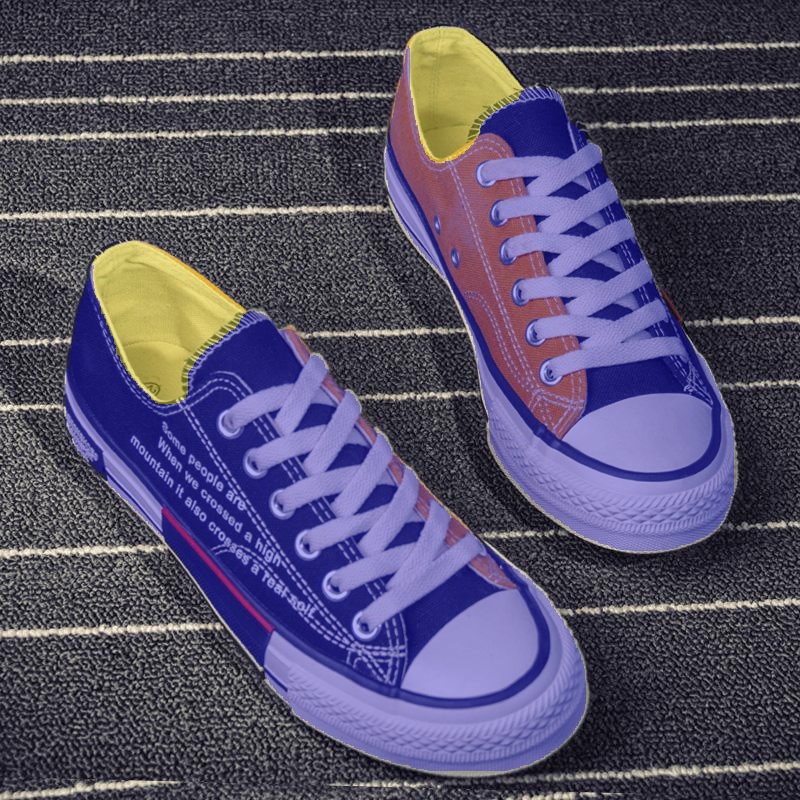}
    \caption{Example of unworn shoe image.}
    \label{fig:shoe unworn}
  \end{subfigure}
  \begin{subfigure}{0.26\linewidth}
    \includegraphics[width=\linewidth]{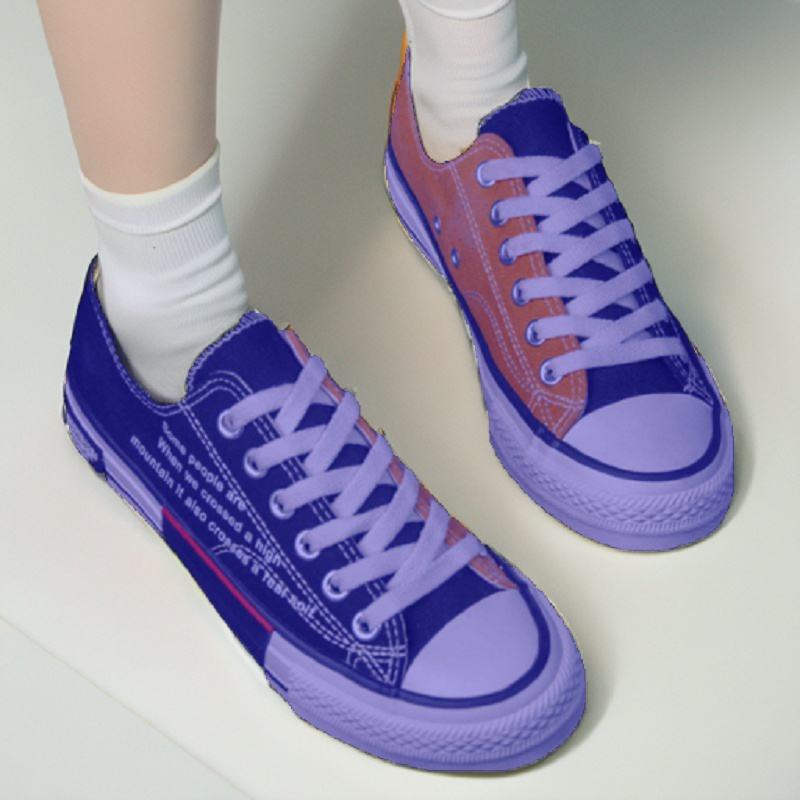}
    \caption{Example of worn shoe image.}
    \label{fig:shoe worn}
  \end{subfigure}
  \caption{\small{Visualization of the visible area and the wearable area. The visible area (shown in \textcolor{blue}{blue}) can be seen in both (\ref{fig:shoe unworn}) and (\ref{fig:shoe worn}), while the wearable area (shown in \textcolor{yellow}{yellow}) only can be seen in (\ref{fig:shoe unworn}).}}
  \label{fig:shoe wearable area example}
\end{figure}

As shown in Figure \ref{fig:shoe wearable area example}, shoes that with/without interactions with human are totally different in visual. Specifically, for images of shoes being worn by human, some part regions of the shoes will be obscured by the legs. Thus, in our shoe-wearing image synthesis task, a pre-process for the given shoe image is necessary to avoid the influence of these obscured regions. Considering the interaction region between the given shoe and human-model, we propose a shoe wearable-area detection module (WD) to detect areas of the given shoe image that may be obscured. However, since the task of occluded shoe part detection has not been addressed in previous literature, this section begins by describing the shoe wearable-area detection task. Subsequently, a detailed solution to this task is presented.

\textbf{Definition of wearable-area detection task.}
\label{subsubsec:shoe wearable task}
The task of shoe wearable-area detection involves distinguishing the visible area and the wearable area in a shoe image. Thus, a shoe image can be divided into three regions: the background, the visible area, and the wearable area. By comparing the differences between an unworn shoe image and its corresponding image being worn by human, the visible area represents the region visible in both images, while the wearable area refers to the region seen only in the unworn image and becomes occluded by the human legs in the worn image. Figure \ref{fig:shoe wearable area example} shows the visualization of the visible area and the wearable area, with the visible area depicted in blue and the wearable area depicted in yellow. In Figure \ref{fig:shoe unworn}, both the visible area and the wearable area can be seen. In Figure \ref{fig:shoe worn}, only the visible area can be seen. The wearable area is occluded by the human leg.

The shoe wearable-area detection task detects the difference area between shoe worn image and shoe unworn image. It gives chances of allowing for the interaction between shoes and humans (i.e., human wears shoes) while maintaining the consistency of shoe identities, which is very important in e-commerce marketing scenario.

\textbf{Detailed Solution: }
\label{subsubsec:shoe wearable solution}
The shoe wearable-area detection task can be treated as the segmentation. Following the task definition, a shoe image can be classified into three classes: background, visible area, and wearable area. And we employ Segformer \cite{xie2021segformer} to implement this task. Since there is currently no existing dataset available for this task, we have constructed a dataset specifically for shoe wearable area detection. The detail of this dataset can be found in Section \ref{subsec:shoe wearable dataset}.

As shown in Figure \ref{fig:pipline}, we first use SAM \cite{kirillov2023segment} to obtain the desired shoe region image $X_{m}$ with black background, then follow the original model architecture of Segformer \cite{xie2021segformer} and set the number of output channel to 3. To improve the generalization of model, we employ several efficient data augmentation methods in the training process to avoid over-fitting. They consist of random Mosic data augmentation \cite{bochkovskiy2020yolov4} with the probability of 0.8, random rotation with the rotation degree in the range of [-15, 15] and the probability of 0.5, and color augmentation, as well as scale augmentation. Finally, after training, the wearable-shoe image $X_{w}$ can be obtained by combining $X_{m}$ with the detected visible areas.


\subsection{Leg-pose Synthesis Module}
\label{subsec:leg pose}
\begin{wrapfigure}{r}{0.5\linewidth}
  \centering\vspace{-2em}
  \begin{subfigure}{0.48\linewidth}
    \includegraphics[width=\linewidth]{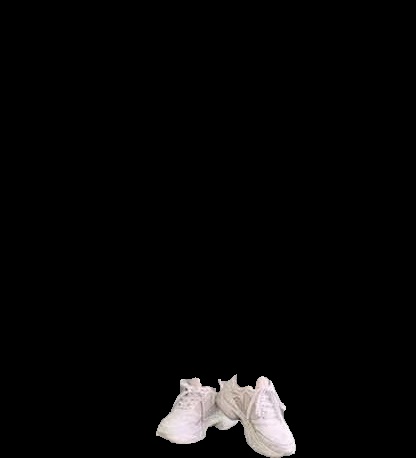}
    \caption{Example of input wearable-shoe $X_{w}$.}
    \label{fig:leg pose eg shoe}
  \end{subfigure}
  \begin{subfigure}{0.48\linewidth}
    \includegraphics[width=\linewidth]{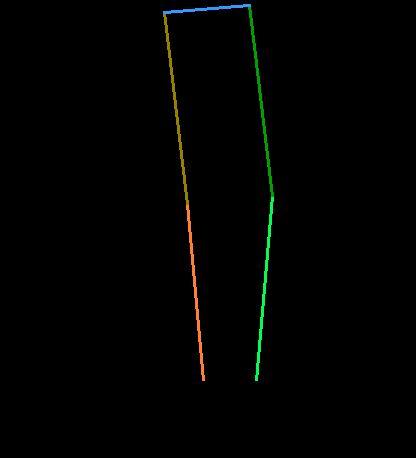}
    \caption{Example of target leg pose image $X_{p}$.}
    \label{fig:leg pose eg label}
  \end{subfigure}
  \caption{Example of leg-pose synthesis task.}
  \label{fig:leg pose task}
\end{wrapfigure}
In this section, we first describe the leg-pose synthesis task. Then, the proposed solution, \ie, leg-pose synthesis (LpS) module,  is introduced.

\textbf{Definition of leg-pose synthesis task.}
\label{subsubsec:leg pose task}
We define the leg-pose synthesis task as generating leg poses from a wearable shoe image, as illustrated in Figure \ref{fig:leg pose task}. Different from the conventional pose estimation task like ViTPose \cite{xu2022vitpose}, the leg-pose synthesis task is a generation task that begins with a provided wearable-shoe image without the accompanying human body structure, and proceeds to generate various potential and reasonable leg poses, \ie, 6 keypoints.

Given a wearable-shoe image, how to generate reasonable leg-poses is challenging. While, considering the fact that the given shoes restrict the ankle position into a determined point and constrain the calf direction into a small range. The calf direction and the shoe direction then jointly constrain the thigh direction into a reasonable range. Therefore, the leg-pose synthesis process must consider not only the orientation of the given shoe, but also the whole structure of the human body, so that the synthesized leg poses are reasonable.

\textbf{Detailed solution: }
\label{subsubsec:leg pose solution}
Initially, we attempted to generate the leg pose image $X_{p}$ by directly conditioning it on the image $X_{w}$, similar to the approach used in LDM \cite{rombach2022high} or ControlNet \cite{zhang2023adding}. However, we encountered challenges with convergence to an optimal state, resulting in generated images with numerous random skeletons and large shadow regions. We conjecture it is because that the original LDM architecture is specially designed for nature images, whereas leg-pose images are more abstract and contain extremely sparse contents. And \textbf{\emph{directly transforming dense real images $X_{w}$ to abstract sparse images $X_{p}$ are difficult}}. To this end, we resort to a two-phase synthesis solution, where we first transform dense nature images to deterministic abstract sparse points, and then convert these deterministic points to various potential points in sparse manner.

Specifically, this two-phase solution contains: (1) deterministic foot keypoints prediction from $X_{w}$; (2) leg-pose synthesis conditioned on these foot points. The system is shown in Figure \ref{fig:pipline}, please refer to LpS module.

In\textbf{ phase-1} of the deterministic foot keypoints estimation, we estimate foot keypoints from a wearable-shoe image $X_{w}$ as shown in Figure \ref{fig:pipline}. It is obvious that the foot keypoints is determined since the foot points are all located on the given shoes. We set 4 keypoints for each foot and 8 keypoints in total for two feet. We implement the widely used pose estimation method Vitpose \cite{xu2022vitpose} to estimate the foot keypoints. The foot keypoints is represented as multi-channel heatmap via MSRA heatmap algorithm \cite{xiao2018simple}. Since we start from the given shoe image and the foot pose estimation phase only estimates 8 foot keypoints, the bounding box of human is not necessary in our scenario. Thus, different from original Vitpose, we remove the input of human detection bounding box and only use the whole image with a fixed size (256, 256) as the input of the model. The data augmentation methods are also different. We implement [random rotation with [-15, 15] degree range and 0.5 probability, rescale and random crop, padding to (256, 256) image size] data augmentation methods instead of the original ones. Then phase-1 can correctly estimate the corresponding 8 foot keypoints from the given shoe image. For latter use, the heatmap information will be rendered to coordinate information, \ie, $C_{p}$.

In \textbf{phase-2} of the leg-pose synthesis, we synthesize reasonable 6 leg-keypoints $L_{0}^{'}$ with the condition of 8 foot-keypoints $C_{p}$ as shown in Figure \ref{fig:pipline}. To model the sparse keypoints information, our leg-pose synthesis phase generates leg poses in [x, y] coordinates instead of heatmaps which are common in 2D pose estimation. The target leg keypoints and the conditional foot keypoints are all normalized into [0, 1] with the formulation $P_{norm} = P / s * 0.9 + 0.1$ where $P_{norm}$ is the normalized coordinate, $P$ is the original coordinate, $s$ is the image size. Meanwhile, $P_{norm}=0$ when the coordinate does not exist \emph{e.g.} only one foot exists in the image. We build a diffusion model with MLPs and optimize with the simplified loss function of \cite{ho2020denoising} denoting as:
\begin{equation}
    L_{lps}=\mathbb{E}_{t,\boldsymbol{L_0},\epsilon}\left[\left\|\epsilon-\epsilon_\theta\left(\sqrt{\overline{\alpha}_t}\boldsymbol{L_0}+\sqrt{1-\overline{\alpha}_t}\epsilon,t,\boldsymbol{C_p}\right)\right\|^2\right]
\end{equation}
where $t\in[0,T]$ is the sampled time steps, $\boldsymbol{C_p}$ is the foot keypoints condition, $\epsilon \sim \mathcal{N}(0,I)$. $\bar{\alpha_{t}}$ are hyper-parameter that controls the added noise. We also use random rotation with [-15, 15] degree range and 0.5 probability to improve the performance. After obtaining the coordinates of leg keypoints $L_{0}^{'}$, we concatenate it with the deterministic foot keypoints $C_{p}$ and render them to the final pose image $X_{p}$.

In addition, since there are currently no existing datasets that provide pairs of shoe images with corresponding foot and leg keypoints, we have created a shoe-pose dataset that includes such pairs. The details of this dataset are provided in Section \ref{subsec:leg-pose dataset}.

\textbf{Remark: }
\label{subsubsec:leg pose summary}
The leg-pose synthesis module is the most import part within our system, it generates a variety of plausible leg poses that align with the given shoe image. These synthesized leg poses serve as reasonable pose constraints for the subsequent human body generation process, ensuring the generation of stable and realistic results.

\subsection{Shoe-wearing Module}
\label{subsec:shoe wearing generation}
After obtaining both $X_{w}$ and $X_{p}$, we build a Shoe-wearing Module to learn to generate hyper-realistic human images wearing the given shoes. As shown in Figure \ref{fig:pipline}, $X_{w}$ and $X_{p}$ are concatenated and then inputted into a ShoeControlNet. The text condition $C_{te}$ is also supplied to both this ShoeControlNet and the frozen base Unet \cite{rombach2022high,zhang2023adding}. Then the overall SW module can be optimized with the simplified loss function of \cite{ho2020denoising} denoted as:
\begin{small}
\begin{equation}\label{eq_sw}
  L_{sw}=\mathbb{E}_{t,\boldsymbol{X_0},\epsilon} \|\epsilon-\epsilon_\omega\left(\sqrt{\overline{\lambda}_t}\boldsymbol{X_0}+\sqrt{1-\overline{\lambda}_t}\epsilon,t,\boldsymbol{X_w,X_p,C_{te}}\right)\|^{2}
\end{equation}
\end{small}
where $t\in[0,T]$ is the sampled time steps; $\boldsymbol{X_w,X_p,C_{te}}$ are conditions of wearable-shoe image, pose image and text embeddings, respectively; $\epsilon \sim \mathcal{N}(0,I)$. $\bar{\lambda_{t}}$ are hyper-parameter that controls the added noise. To train this module, we also employ our collected shoe-leg dataset which will be introduced in Section \ref{subsec:leg-pose dataset}. At inference time, in order to keep the ID-consistency of shoes, $X_{w}$ will be copied and then pasted back to the generated image. 
\section{Shoe-wearing Dataset}
\label{sec:dataset}
In Section \ref{sec:method}, we introduced two novel tasks: the shoe wearable-area detection task and the leg-pose synthesis task. As there are currently no suitable existing datasets available for these tasks, we have constructed a shoe-wearing dataset specifically for the proposed \emph{ShoeModel} system. This dataset comprises two parts: the shoe wearable-area detection dataset and the shoe-leg dataset. These datasets facilitate the training of WD and LpS/SW, respectively. Detailed information regarding these datasets is provided in the following subsections.

\subsection{Wearable-area detection dataset}
\label{subsec:shoe wearable dataset}

Table \ref{tab:shoe wearable dataset example} presents examples of our wearable-area detection dataset. In the annotations, the black area represents the background, while the blue and yellow areas indicate the visible area and the wearable area, respectively. We annotate 2,051 images in total with the help of SAM \cite{kirillov2023segment}. This well annotated dataset supports the training of WD module.
\begin{table}[t!]
    \centering
    \footnotesize
    \begin{minipage}[t]{0.4\linewidth}
      \begin{tabular}{|cc|cc|}
        \hline
        \shortstack{\bf Shoe \\ \bf Image} & \shortstack{\bf Area \\ \bf Annot.} & \shortstack{\bf Shoe \\ \bf Image} & \shortstack{\bf Area \\ \bf Annot.} \\
        \hline
        \hline
        \includegraphics[width=0.11\linewidth]{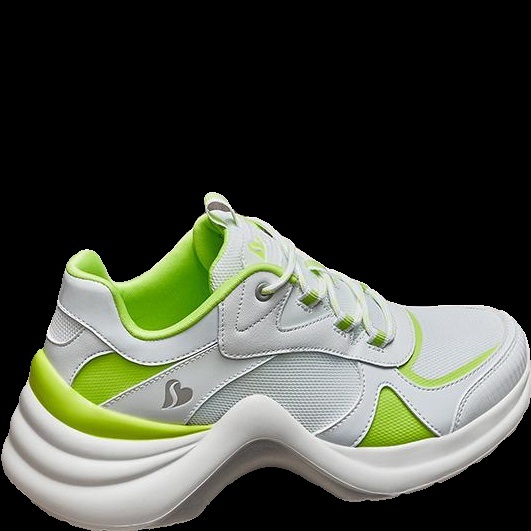} & \includegraphics[width=0.11\linewidth]{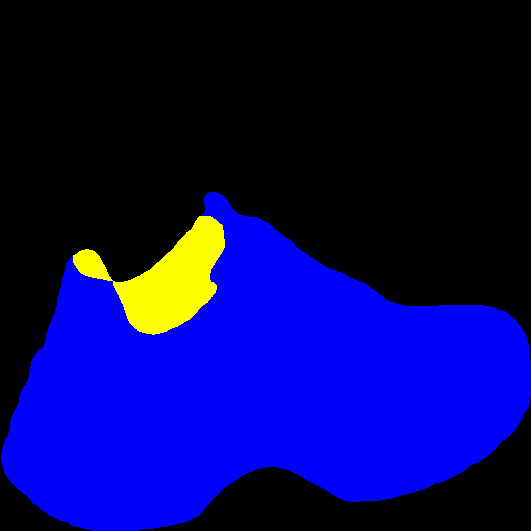} & \includegraphics[width=0.11\linewidth]{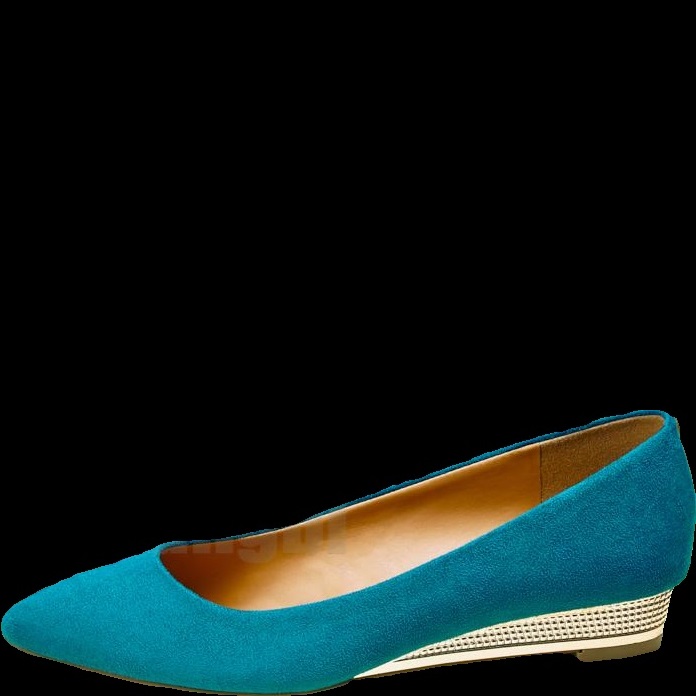} & \includegraphics[width=0.11\linewidth]{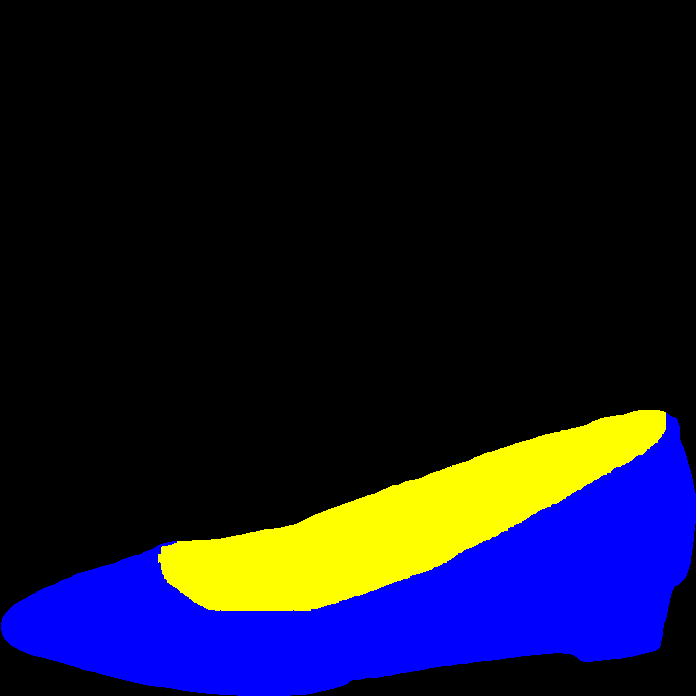}\\
        \includegraphics[width=0.11\linewidth]{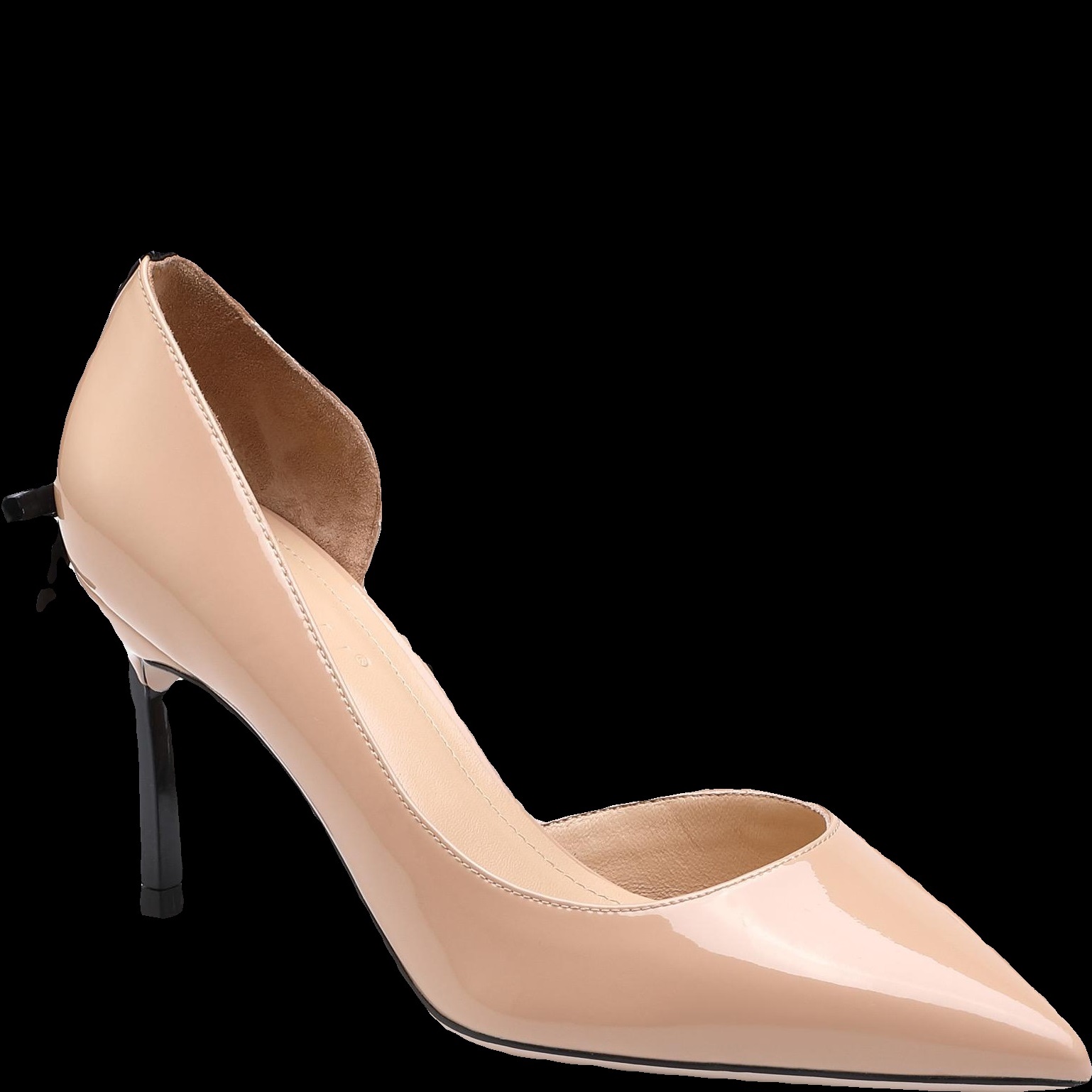} & \includegraphics[width=0.11\linewidth]{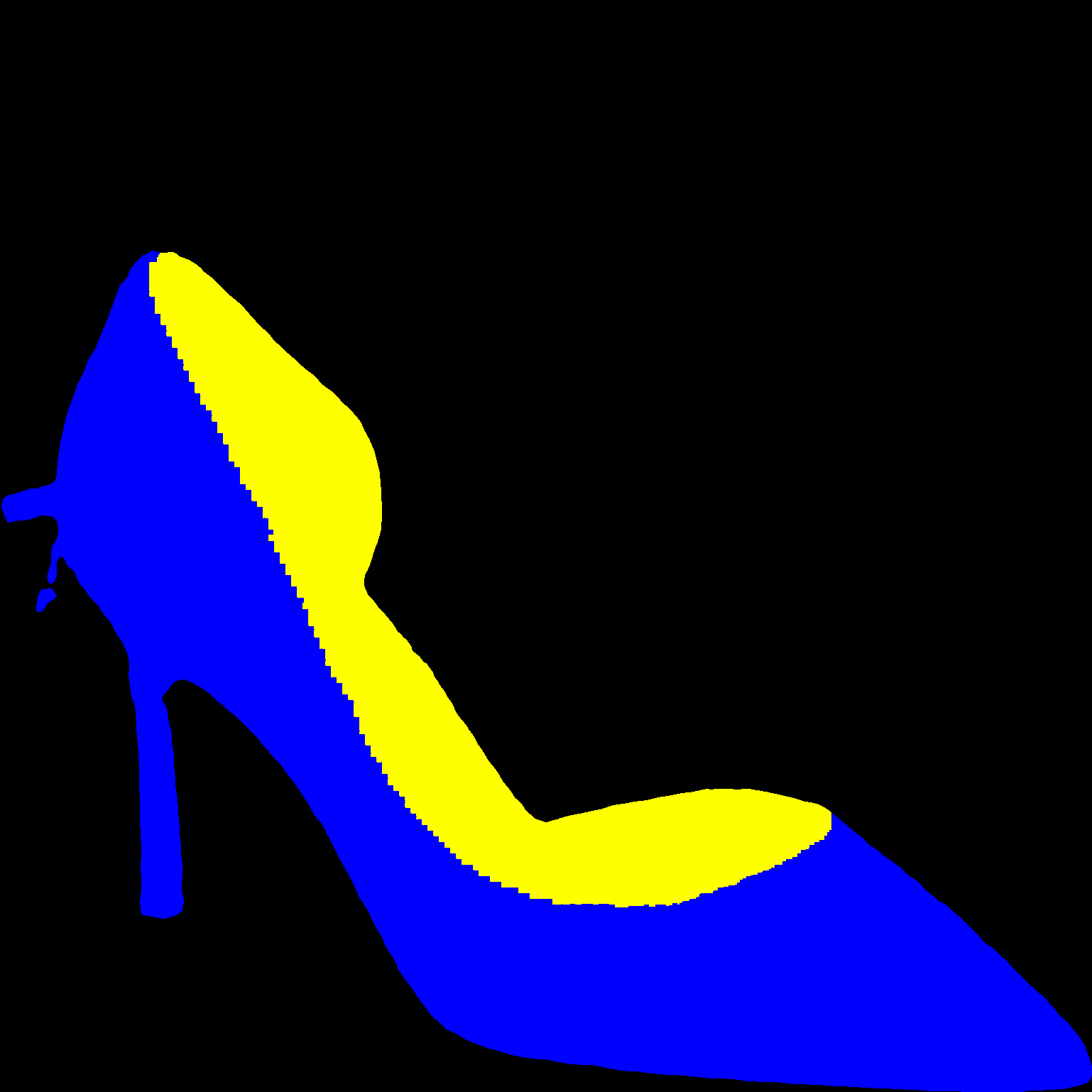} & \includegraphics[width=0.11\linewidth]{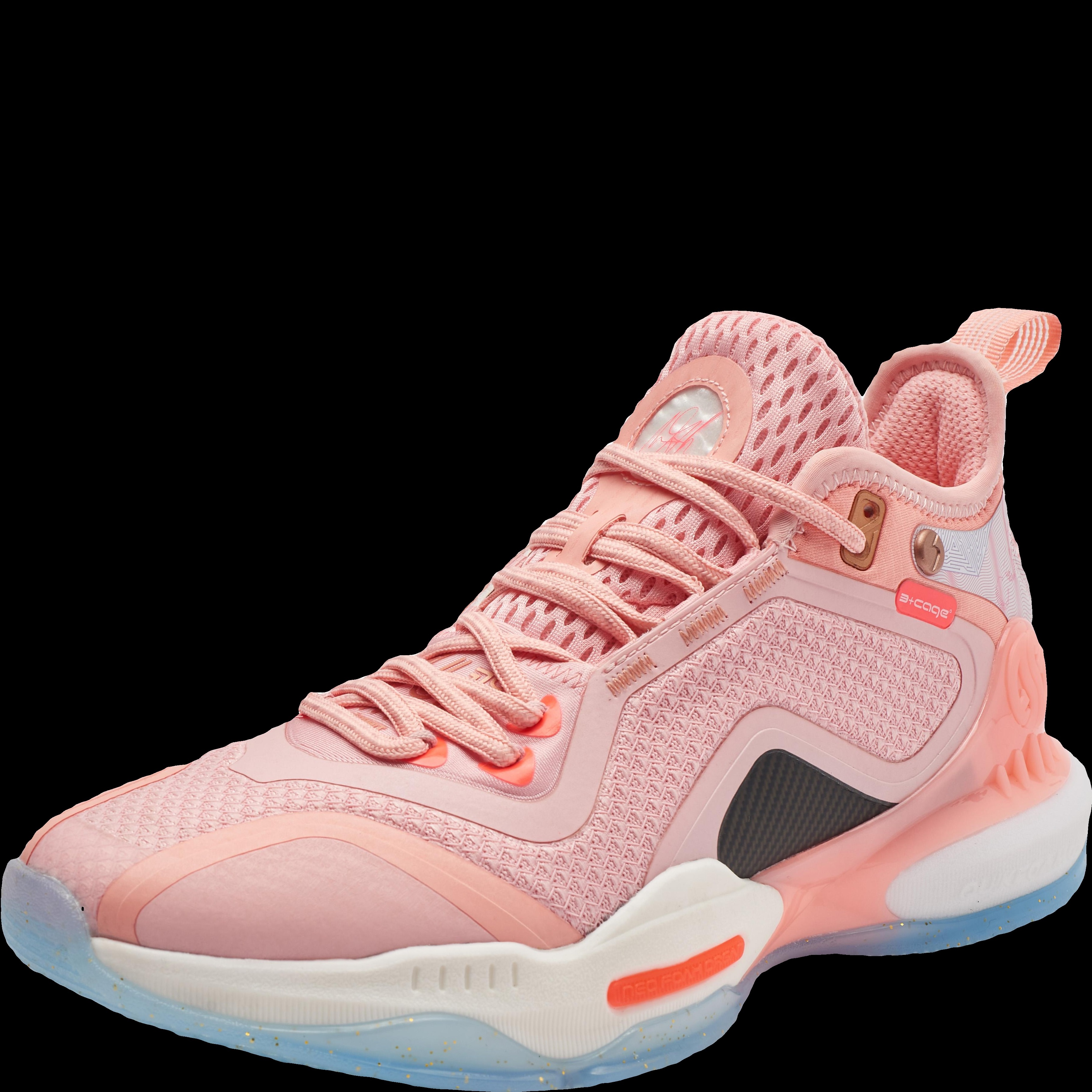} & \includegraphics[width=0.11\linewidth]{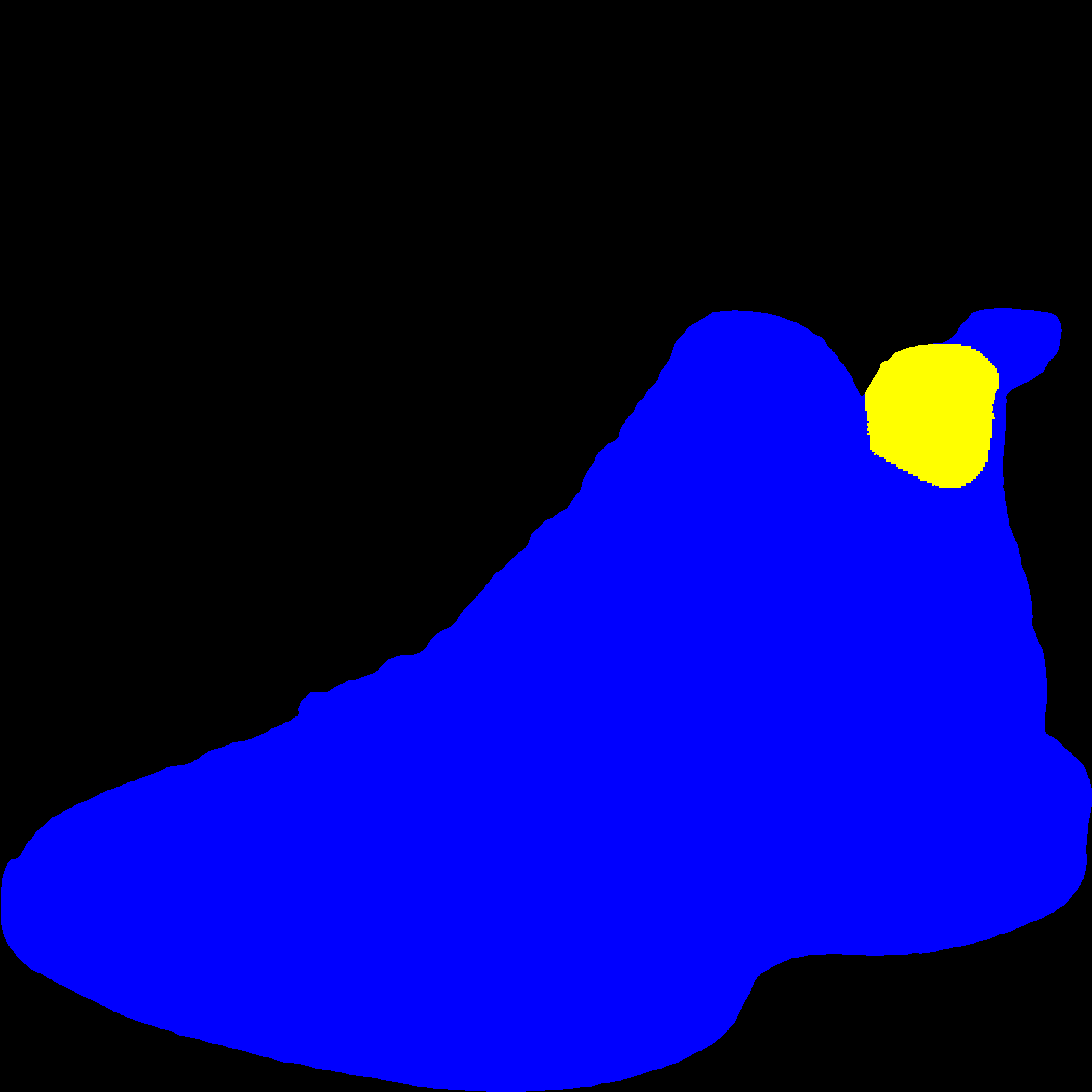}\\
        \includegraphics[width=0.11\linewidth]{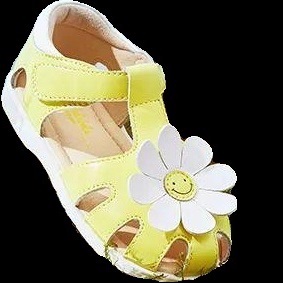} & \includegraphics[width=0.11\linewidth]{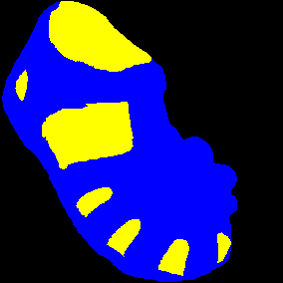} & \includegraphics[width=0.11\linewidth]{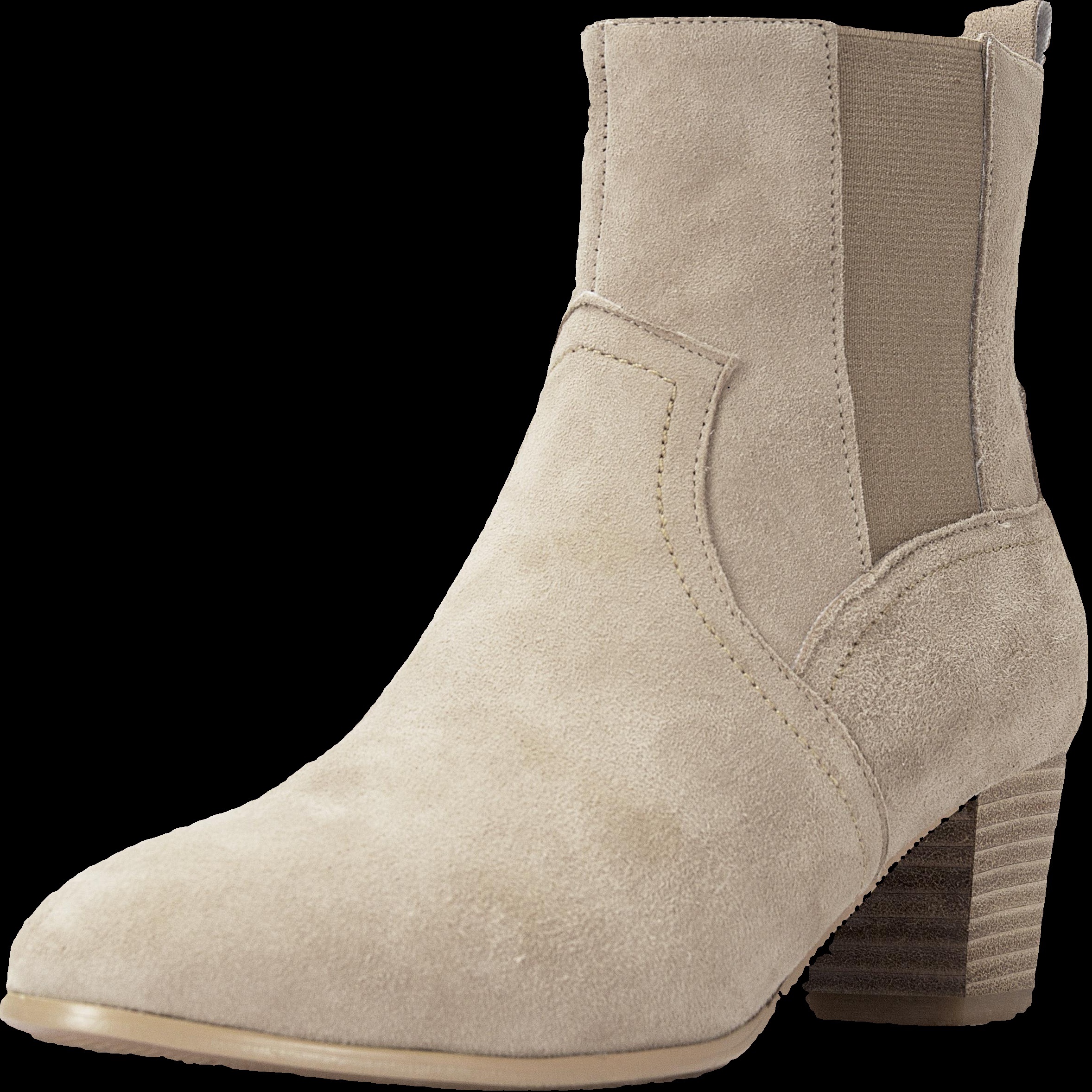} & \includegraphics[width=0.11\linewidth]{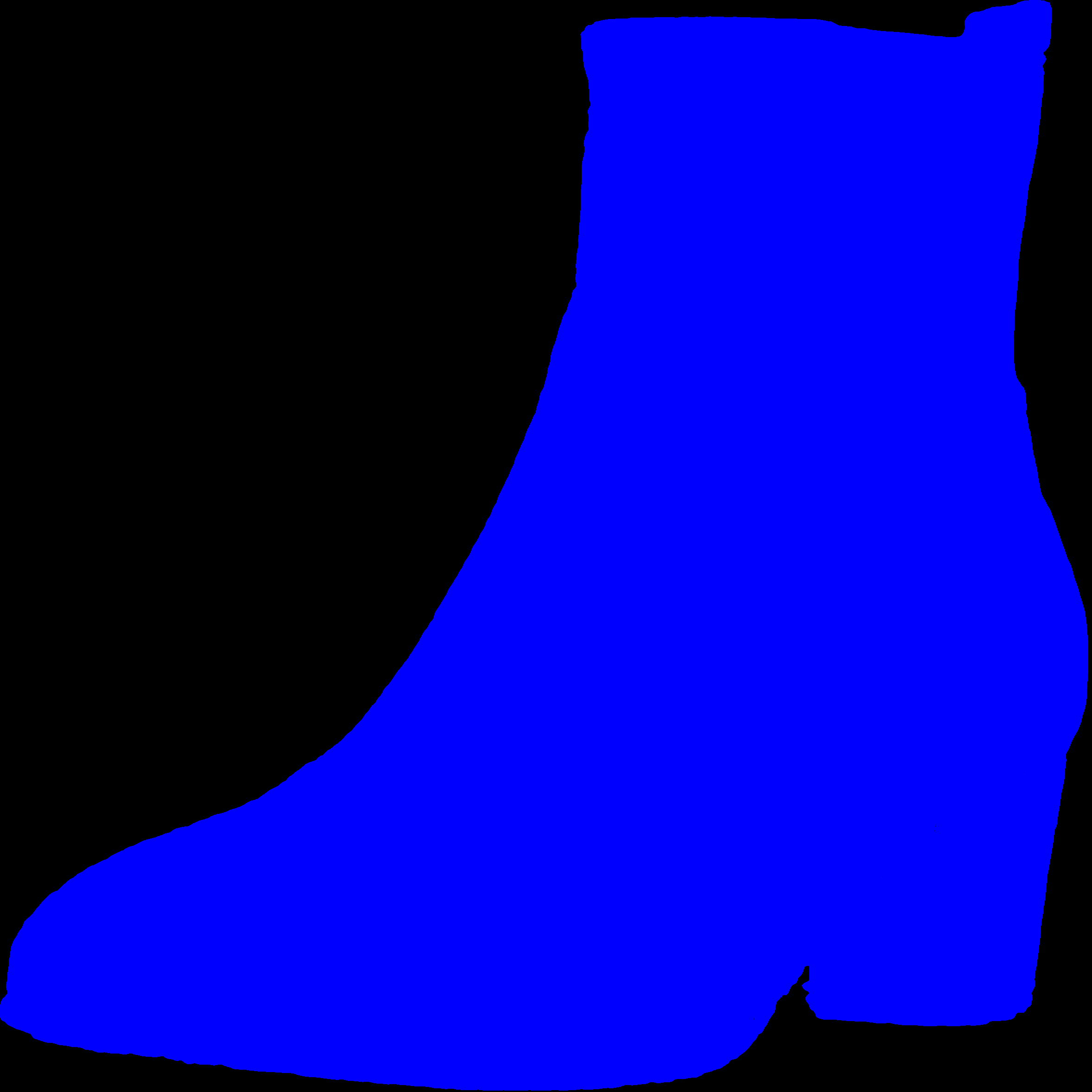}\\
        \includegraphics[width=0.11\linewidth]{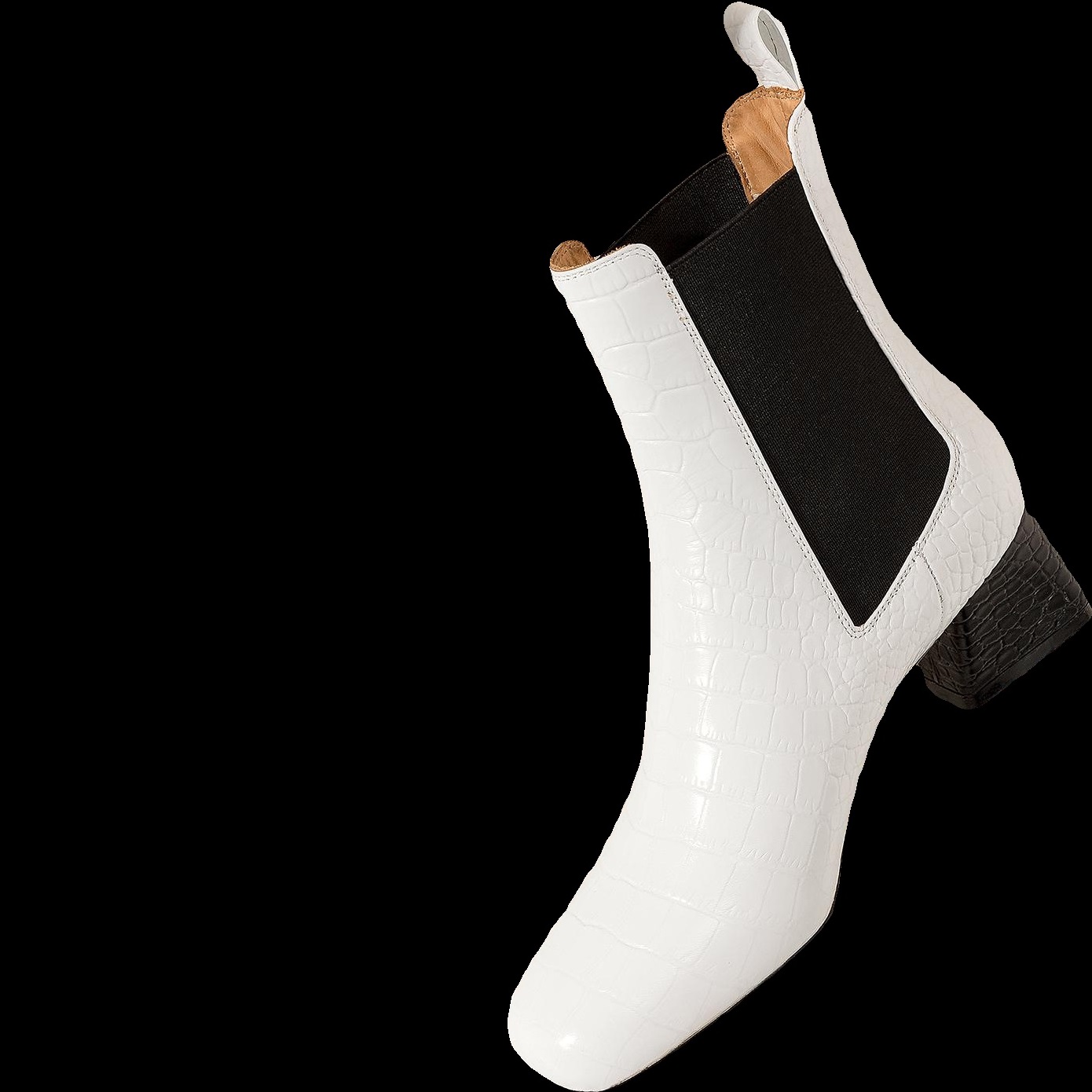} & \includegraphics[width=0.11\linewidth]{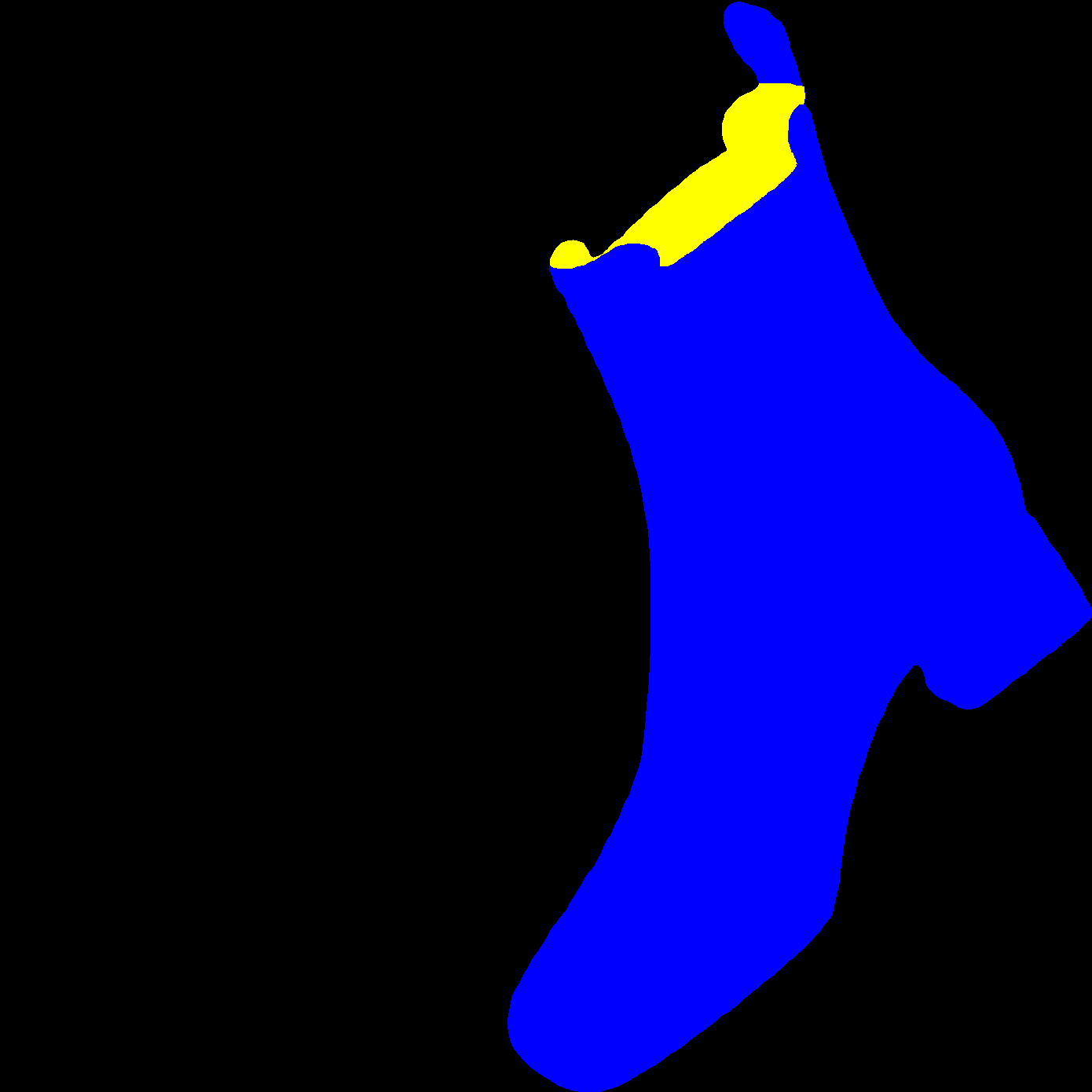} & \includegraphics[width=0.11\linewidth]{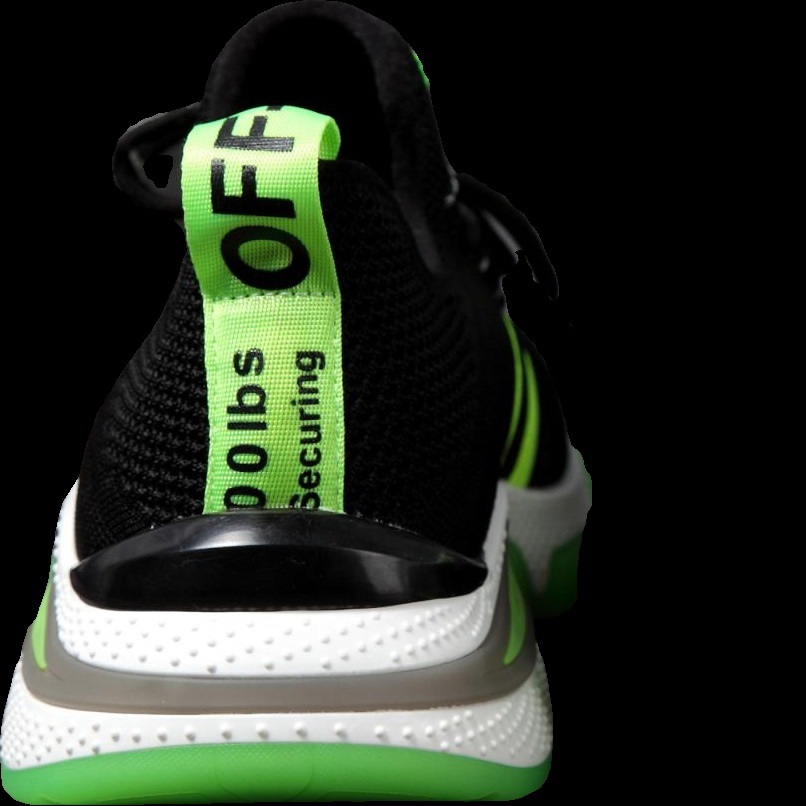} & \includegraphics[width=0.11\linewidth]{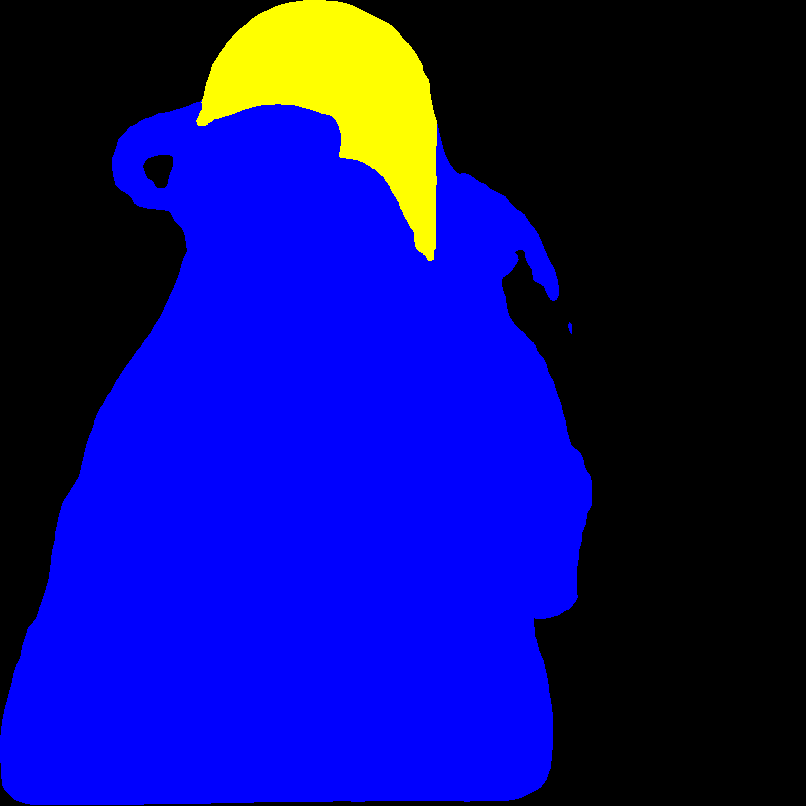}\\
        \hline
    \end{tabular}
    \caption{\small{Examples of shoe wearable-area detection dataset. In the annotations, the background is shown in \textbf{\textcolor{black}{black}}, the visible area is shown in \textbf{\textcolor{blue}{blue}}, and the wearable area is shown in \textbf{\textcolor{yellow}{yellow}}}.}
    \label{tab:shoe wearable dataset example}
    \end{minipage}
    \begin{minipage}[t]{0.49\linewidth}
      \begin{tabular}{|ccc|ccc|}
        \hline
        \shortstack{\bf Raw \\ \bf Image} & \shortstack{\bf Shoe \\ \bf Image} & \shortstack{\bf Pose \\ \bf Annot.} & \shortstack{\bf Raw \\ \bf Image} & \shortstack{\bf Shoe \\ \bf Image} & \shortstack{\bf Pose \\ \bf Annot.} \\
        \hline
        \hline
        \includegraphics[width=0.06\linewidth]{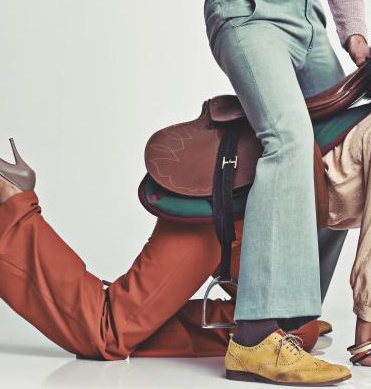} &         \includegraphics[width=0.06\linewidth]{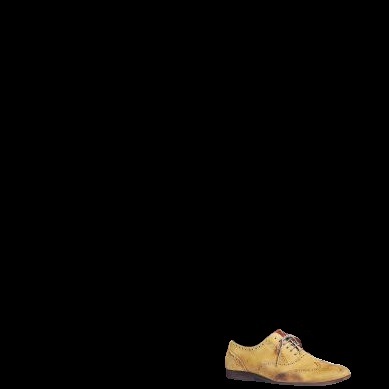} & \includegraphics[width=0.06\linewidth]{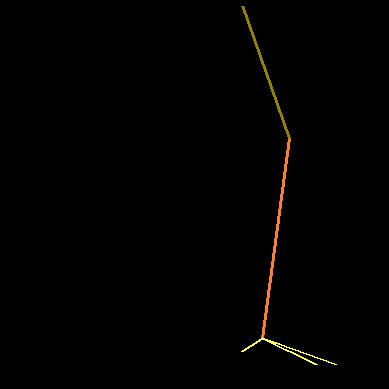} & \includegraphics[width=0.06\linewidth]{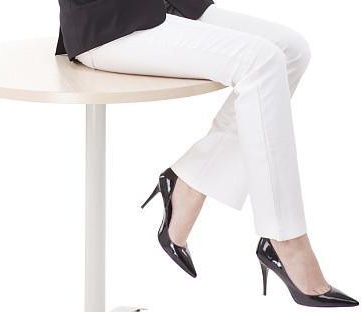} & \includegraphics[width=0.06\linewidth]{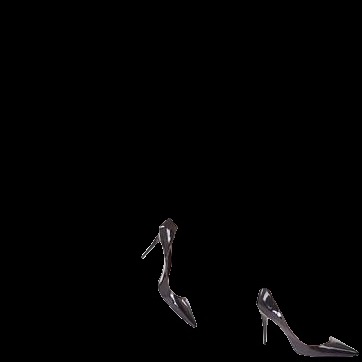} & \includegraphics[width=0.06\linewidth]{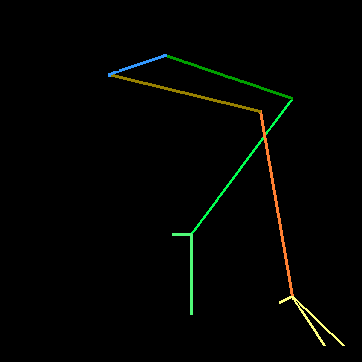}\\
        \includegraphics[width=0.06\linewidth]{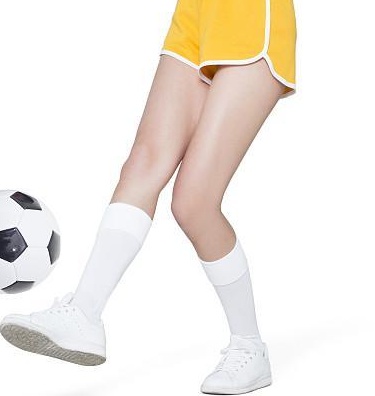} & \includegraphics[width=0.06\linewidth]{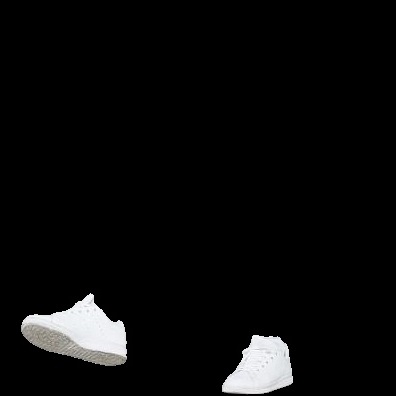} & \includegraphics[width=0.06\linewidth]{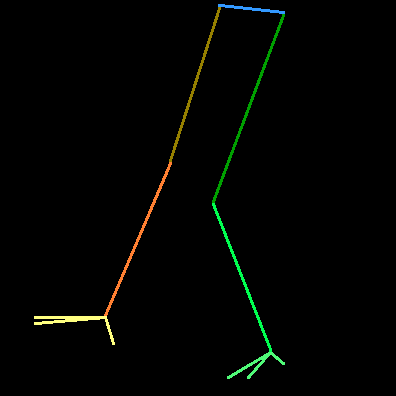} & \includegraphics[width=0.06\linewidth]{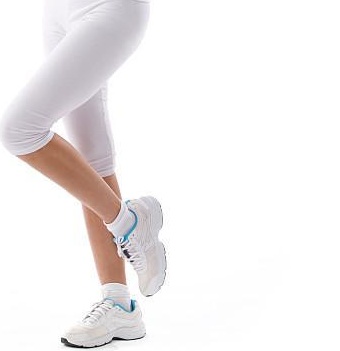} & \includegraphics[width=0.06\linewidth]{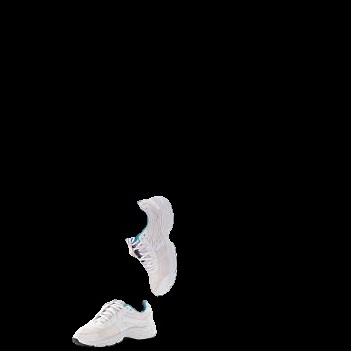} & \includegraphics[width=0.06\linewidth]{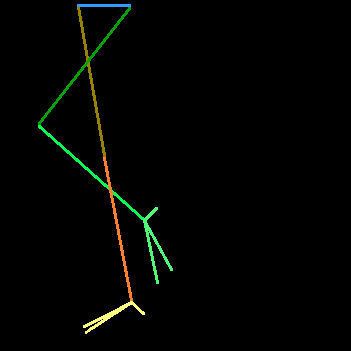}\\
        \includegraphics[width=0.06\linewidth]{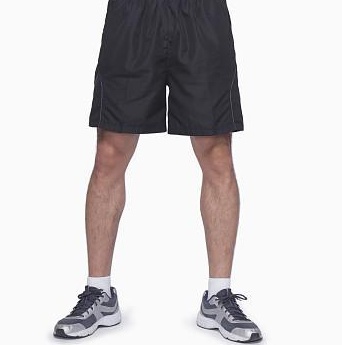} & \includegraphics[width=0.06\linewidth]{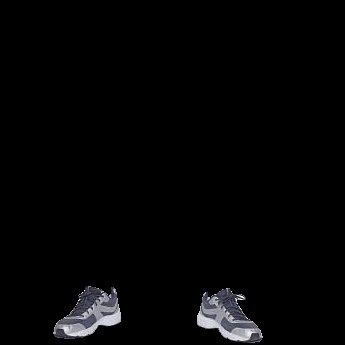} & \includegraphics[width=0.06\linewidth]{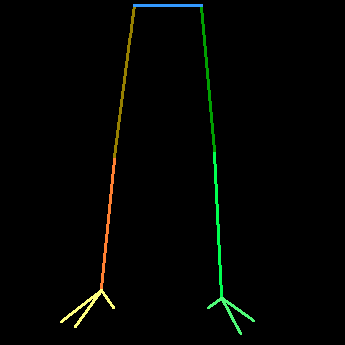} & \includegraphics[width=0.06\linewidth]{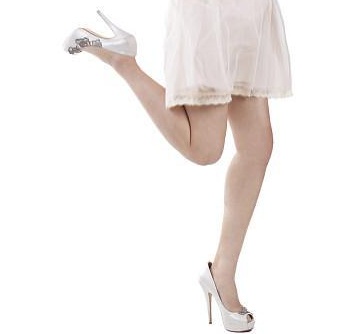} & \includegraphics[width=0.06\linewidth]{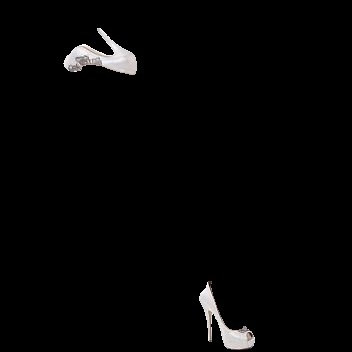} & \includegraphics[width=0.06\linewidth]{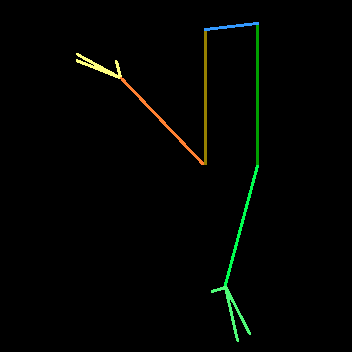}\\
        \includegraphics[width=0.06\linewidth]{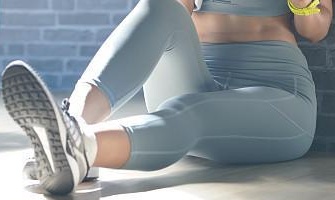} & \includegraphics[width=0.06\linewidth]{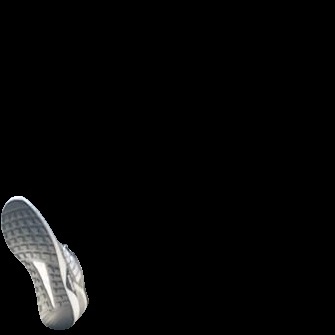} & \includegraphics[width=0.06\linewidth]{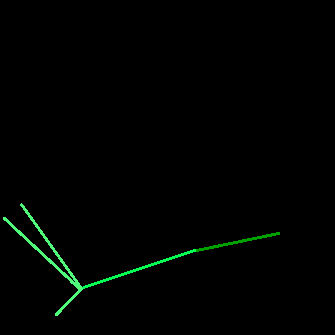} & \includegraphics[width=0.06\linewidth]{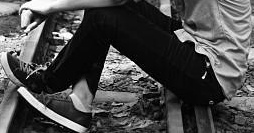} & \includegraphics[width=0.06\linewidth]{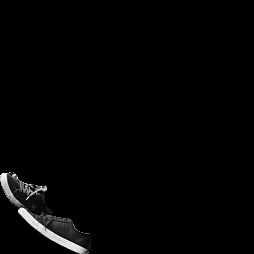} & \includegraphics[width=0.06\linewidth]{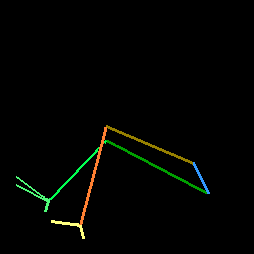}\\
        \hline
    \end{tabular}
    \caption{Examples of shoe-leg dataset.}
    \label{tab:leg pose dataset example}
    \end{minipage}
\end{table} 
\subsection{Shoe-leg dataset}
\label{subsec:leg-pose dataset}
Although numerous 2D and 3D whole-body human datasets have been proposed in recent years, such as the COCO wholebody dataset \cite{jin2020whole} and Human3.6M wholebody dataset \cite{zhu2023h3wb}, they do not fulfill our specific requirements for generating hyper-realistic advertising images featuring shoes worn by humans. To this end, we construct a shoe-leg dataset comprising tuples of raw images, shoe-only images and pose skeleton annotations. We first collect millions of human images from internal advertising images, and use a pre-trained YOLOX-l model \cite{ge2021yolox} to filter the human images. The wholebody poses are then detected through the ViTPose model \cite{xu2022vitpose}. The leg poses and foot poses in the detection results are combined to formulate the 12-keypoint annotations. Additionally, we use SAM \cite{kirillov2023segment} for shoe segmentation. Finally, we obtain 669,808 images in total to construct the shoe-leg dataset. Table \ref{tab:leg pose dataset example} shows the paired examples of this dataset.
\section{Experiments}
\label{sec:experiments}
\textbf{Implementation details:} For WD module, we train it for 160k iterations with learning rate of 6e-5 and with batchsize of 16 on 4 NVIDIA-V100 GPUs. The images are resized to 640$\times$640. For LpS phase-1 stage, we train it for 180 epochs with learning rate of 5e-4 and with batchsize of 240. The images are resized to 256$\times$256. For LpS phase-2 stage, we train it for 700k iterations with learning rate 1e-4 and with batchsize of 4096. For SW module, we train it for 500k iterations with learning rate of 1e-5 and with batchsize of 128. The images are randomly cropped to 512$\times$512. The latter three models are all trained on 8 NVIDIA-V100 GPUs. The time steps $T$ are 25 and 1000 for LpS phase-2 and SW, respectively. DDIM \cite{song2020denoising} sampler is used with CFG 7.0.

\textbf{Evaluation metrics:} We use several metrics to evaluate the performances: (1) Fr$\acute{e}$chet Inception Distance \cite{heusel2017gans} (FID) and Kernel Inception Distance \cite{binkowski2018demystifying} (KID) are used to evaluate the image quality, (2) CLIP score is used to evaluate the text-image consistency. Moreover, we also carry out user studies to evaluate their perceptual plausibility.

For testing the overall system of \emph{ShoeModel}, we collect another 2k test set in which only shoes are shown in the images.
\begin{table*}[t]
  \centering
  \caption{Quantitative comparisons between our proposed \emph{ShoeModel} and other diffusion models (including the Text-to-Image SD1.5 and SDXL methods, the Condition-to-Image SD1.5-inpaint and ControlNet-inpaint methods) on our 2k test set. We also provide the Competency Matrix comparisons to show the differences between \emph{ShoeModel} and other methods.}
  \resizebox{0.8\linewidth}{!}{
    \begin{tabular}{lcccccc}
    \hline
    \multicolumn{1}{c}{\multirow{2}[2]{*}{Method}} & \multicolumn{3}{c}{Metrics} & \multicolumn{2}{c}{Competency Matrix} &  \multicolumn{1}{c}{User-Study}\\
\cmidrule{2-6}          & FID $\downarrow$   & KID$_{\times1k}$ $\downarrow$  & CLIP score $\uparrow$& Specified-Shoe & Quality of Generation & Num $\uparrow$\\
\hline
    SD1.5 \cite{rombach2022high} & 31.62 & 12.06 & 23.84 & Not Support & - &-\\
    SDXL \cite{podell2023sdxl}  & 30.24 & 7.49  & \textbf{24.38} & Not Support & - &-\\
    \hline
    SD1.5-inpaint \cite{rombach2022high} & 30.77 & 11.95 & 23.49 & Partial Support & inferior &54 \\
    ControlNet-inpaint \cite{zhang2023adding} & 29.83 & 11.28 & 23.65 & Partial Support & inferior &56\\
    Paint-by-Example \cite{yang2023paint} & 31.92 & 11.68 & 22.37 & Partial Support & inferior & 30.5\\
    \hline
    ShoeModel (ours) & \textbf{22.54} & \textbf{6.67}  & 24.02 & \textbf{Support} & \textbf{good} & \textbf{1859.5}\\
    \hline
    \end{tabular}%
    }
  \label{tab:quantitative}%
\end{table*}%
\subsection{Quantitative Analysis}
\textbf{To our best knowledge, since this paper is the first work to do this kind of task, \eg, producing hyper-realistic human advertising images interacting with the specified objects and keep the object-ID consistency.} There are few existing works to be compared. While, we still compare our method with several other works, including the Text-to-Image (T2I) SD1.5\cite{rombach2022high} and SDXL \cite{podell2023sdxl} methods, the Condition-to-Image (C2I) SD1.5-inpaint \cite{rombach2022high}, ControlNet-inpaint \cite{zhang2023adding} and Paint-by-Example \cite{yang2023paint} methods. For T2I methods, we use the same prompt to generate the images. For C2I methods, we input both the shoe image and the prompt. From Table \ref{tab:quantitative}, one can observe that our \emph{ShoeModel} outperforms all the other methods in both FID and KID metrics, and achieves second best performance over the CLIP score metric. Note that since SDXL \cite{podell2023sdxl} uses multiple powerful text-encoders, which are with larger Unet models, it achieves the best text-image consistency.

Moreover, we also give the competency matrix comparisons as in Table \ref{tab:quantitative} to show the importance and superiority of our work. And user-study is also conducted by twenty volunteers. The images generated by different methods under the same conditions were randomly shuffled and simultaneously presented to the volunteers, and only the image selected as the winner was considered for evaluation. The averaged number of wins is shown in Table \ref{tab:quantitative}, demonstrating that \emph{ShoeModel} outperforms other methods by a large margin.
\subsection{Qualitative Analysis}\label{sec_qualitative}
\begin{figure}[t]
  \centering
  \includegraphics[width=0.6\linewidth]{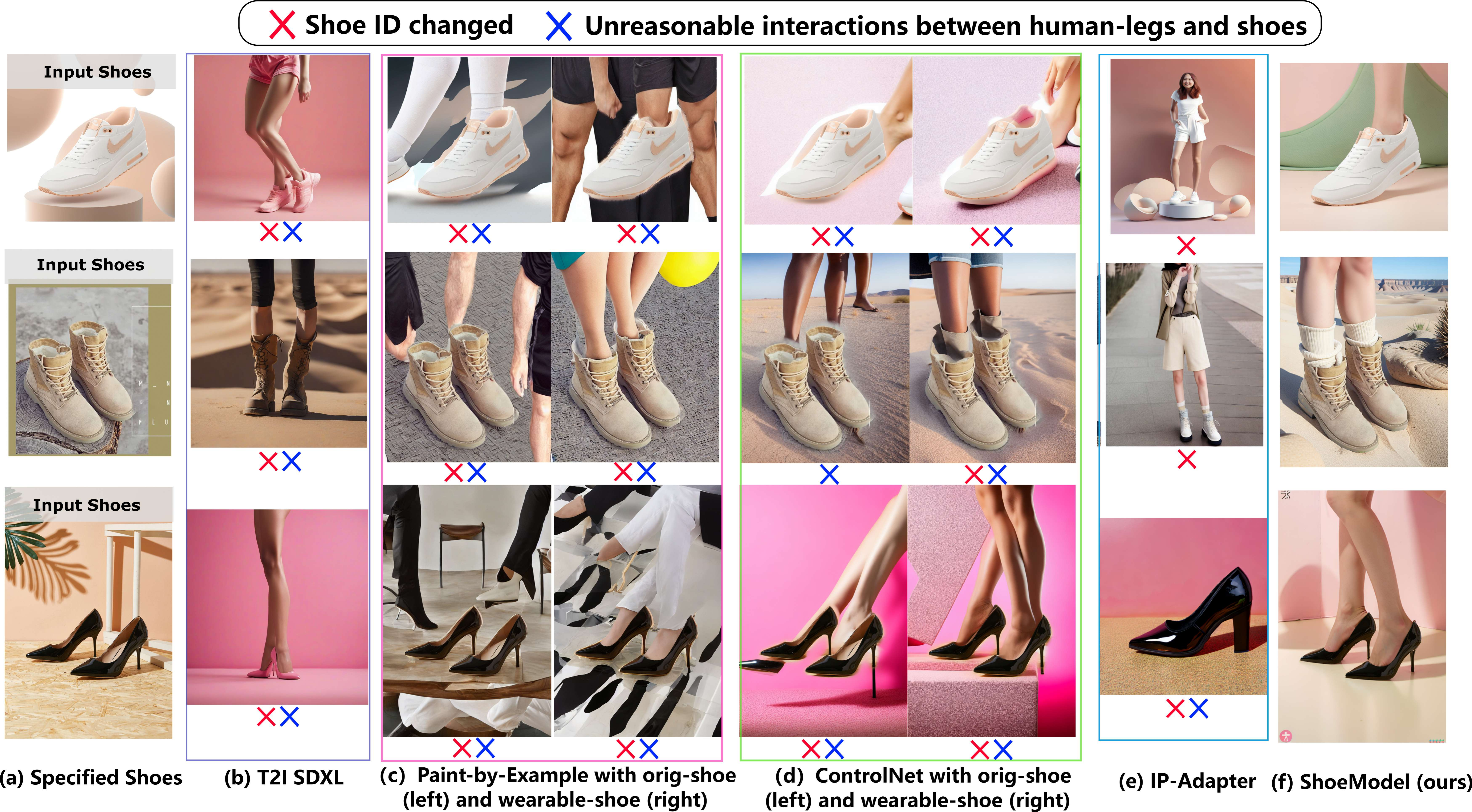}
  \caption{Qualitative comparisons with SDXL \cite{podell2023sdxl}, Paint-by-example \cite{yang2023paint}, ControlNet-Inpainting \cite{zhang2023adding} and Ip-Adapter \cite{ye2023ip-adapter}. One can observe that our \emph{ShoeModel} performs the best in terms of image quality, shoe-ID consistency and reasonability of interactions between human and shoes. More results are in supplementary file.}\label{fig_qualitative}
\end{figure}
Since the quantitative results sometimes diverge lots with human preference, we provide the qualitative comparisons with several representative related works, including text2img-based SDXL \cite{podell2023sdxl}, painting-based Paint-by-example \cite{yang2023paint} and ControlNet-Inpainting \cite{zhang2023adding}, training-free based Ip-Adapter \cite{ye2023ip-adapter}, and training-based Dreambooth \cite{ruiz2023dreambooth}. For fair comparisons, we use the same input shoes and prompts for all methods. From Figure \ref{fig_qualitative}, we observe that the T2I-based SDXL\cite{podell2023sdxl} naturally doesn't support interactions with the given shoes, since it only receives text input.

\textbf{Compared with Painting-based methods:} when provided the complete shoes, as shown in the third and fifth columns of Figure \ref{fig_qualitative}, Paint-by-example \cite{yang2023paint} and ControlNet-Inpainting can use this input to guide the generation, but they struggle to generate reasonable interactions between human-legs and shoes since they cannot handle the wearable-areas of shoes. Moreover, even with artificial assistance in removing the wearable areas of shoes, as shown in the fourth and sixth columns, these two methods still cannot generate satisfactory interactions, as well as often producing new contents of shoes around the given ones (\ie, changing the identity of the specified shoes). We also verify the failure of painting-based idea by removing $X_{p}$ from our ShoeModel in the subsequent ablation section. All these results show the existing painting-based methods are not a good choice for our task.

\textbf{Compared with Training-free methods:} different from \cite{yang2023paint,zhang2023adding}, Ip-Adapter \cite{ye2023ip-adapter} first compresses the input objects into high-level feature embeddings via certain image encoders, and then produces output image conditioned on these image embeddings. However, this kind of method cannot guarantee the ID consistency between input and output. It is because it uses an image encoder (CLIP) to compress the pixel-level shoes into high-level latent-space and then generates images based on this latent-space, the image encoder is pre-trained to mainly focus on the object-level concepts instead of the minor details within objects, such as pixel-level color, logo, texture and so on. As a result, information loss of shoe-detail is inevitable during this compression process, which in turn affects the consistency of the generated shoe-ID. Figure \ref{fig_qualitative}.(e,f) verifies this. 

\begin{figure}[t]
  \centering
  \includegraphics[width=0.8\linewidth]{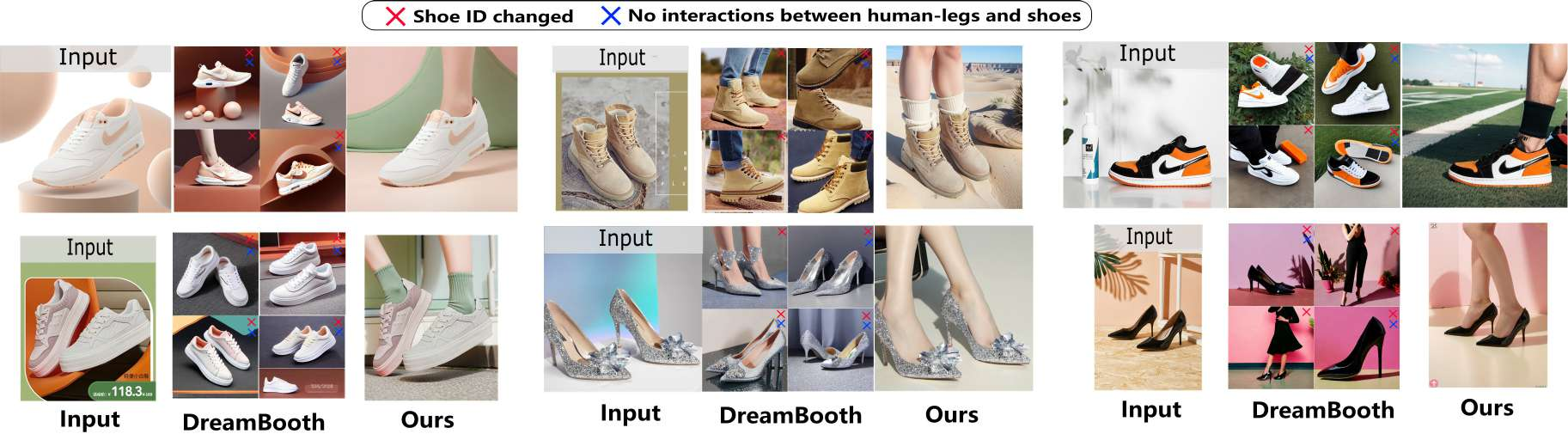}
  \caption{Qualitative comparisons with DreamBooth \cite{ruiz2023dreambooth}. One can observe that our \emph{ShoeModel} performs the best in terms of image quality, shoe-ID consistency and reasonability of interactions between human and shoes.}\label{fig_qualitative_dreambooth}
\end{figure}

\textbf{Compared with Training-based method:} DreamBooth \cite{ruiz2023dreambooth} is a representative method. We conduct experiments as shown in Figure \ref{fig_qualitative_dreambooth}. To improve the training effects, we first augment the single input shoe to 10 counterparts and then train the DreamBooth. From this figure, one can observe that this training-based method still cannot ensure both the shoe identity and the interaction between shoes and human-legs. It is because using small-scale data is hard to change the manifold learned by Diffusion Model, as a result, the details of shoes (like color, logo, text, or shape) cannot be well maintained. Consequently, it is even more challenging to have a human model wear these shoes convincingly.

In summary, images output by these above methods either are with changed shoe IDs or with unreasonable leg-poses, as a result, they cannot be used in actual E-commerce marketing scene(\eg, it may lead to false advertising and lack appeal due to unattractive images, respectively). In contrast, our \emph{ShoeModel} excels in generating hyper-realistic images that exhibit not only reasonable interactions between humans and shoes but also maintain the inherent identity of the shoes unchanged.
\begin{figure}[!h]
  \centering
  \begin{minipage}[t]{0.4\linewidth}
    \includegraphics[width=1\linewidth]{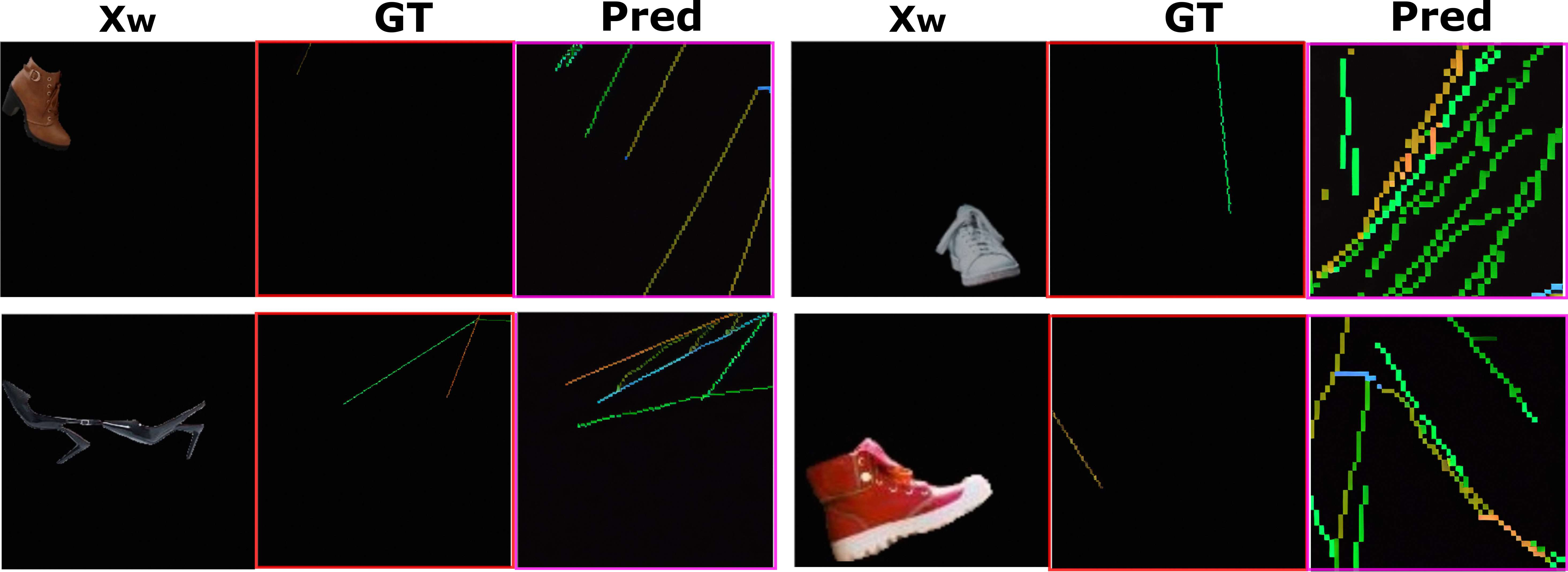}
  \caption{The prediction results of training ControlNet-based LpS module.}\label{fig_lsp_ab}
  \end{minipage}
  \begin{minipage}[t]{0.56\linewidth}
  \resizebox{1\linewidth}{!}{
    \begin{tabular}{c|cc|cc|cc|ccc}
    \hline
    \multirow{2}[2]{*}{} & \multicolumn{2}{c|}{BG} & \multicolumn{2}{c|}{Visable-Area} & \multicolumn{2}{c|}{Wearable-Area} & \multicolumn{3}{c}{All} \\
         & IoU   & Acc   & IoU   & Acc   & IoU   & Acc   & mIoU  & mAcc  & aAcc \\
\hline
    \textbf{WD}    & 99.42 & 99.67 & 97.3  & 98.62 & \textbf{85.55} & \textbf{92.45} & 94.09 & 96.91 & 98.60 \\
    \hline\hline
        & \multicolumn{4}{c|}{Other-Area} & \multicolumn{2}{c|}{Wearable-Area} & \multicolumn{3}{c}{All} \\          & \multicolumn{2}{c}{IoU} & \multicolumn{2}{c|}{Acc} & IoU   & Acc   & mIoU  & mAcc  & aAcc \\
\hline
    WD(2-classes) & \multicolumn{2}{c}{99.45} & \multicolumn{2}{c|}{99.60} & 79.59 & 89.15 & 89.52 & 94.38 & 98.11 \\
    \hline
    \end{tabular}%
    }\captionof{table}{Results of WD module on 200 tested images of Wearable-area detection dataset.}
  \label{tab:wd_res}%
  \end{minipage}
\end{figure}
\subsection{Ablation Study}
\textbf{Ablation study on the design of LpS:} In Section \ref{subsec:leg pose}, we emphasized that directly transforming dense real images $X_{w}$ to the abstract sparse images $X_{p}$ are difficult. To demonstrate that, we train a new leg-pose synthesis model based on the popular used controlnet \cite{zhang2023adding}. In which, the conditions are $X_{w},C_{te}$ and the output is targeted by $X_{p}$. Figure \ref{fig_lsp_ab} shows the example results. We can observe that when given the wearable-shoe image $X_{w}$, the prediction $X_{p}$ is very noisy and thus cannot be used to guide the subsequent generation.

\textbf{Results of WD module:} Since the precision of wearable-area detection module is very important for the final performances of the overall system, we evaluate the segmentation results of WD module as in Table \ref{tab:wd_res}. From this table, we can observe that our WD module is indeed accurate and thus can be used to provide precise $X_{w}$ image conditions for the subsequent generation stages. Moreover, we also conduct an ablation study by setting only two target classes, \ie, wearable-area and other area. Performances can also be found in Table \ref{tab:wd_res}, denoted as WD(2-classes), one can observe that this strategy is worse, demonstrating that our 3-classes fine-grained annotations are indeed useful and good for accurate segmentation.

\begin{table}[!t]
  \centering
  \caption{Results of WD module on 200 tested images of Wearable-area detection dataset.}
  \resizebox{0.5\linewidth}{!}{
    \begin{tabular}{c|cc|cc|cc|ccc}
    \hline
    \multirow{2}[2]{*}{} & \multicolumn{2}{c|}{BG} & \multicolumn{2}{c|}{Visable-Area} & \multicolumn{2}{c|}{Wearable-Area} & \multicolumn{3}{c}{All} \\
         & IoU   & Acc   & IoU   & Acc   & IoU   & Acc   & mIoU  & mAcc  & aAcc \\
\hline
    \textbf{WD}    & 99.42 & 99.67 & 97.3  & 98.62 & \textbf{85.55} & \textbf{92.45} & 94.09 & 96.91 & 98.60 \\
    \hline\hline
        & \multicolumn{4}{c|}{Other-Area} & \multicolumn{2}{c|}{Wearable-Area} & \multicolumn{3}{c}{All} \\          & \multicolumn{2}{c}{IoU} & \multicolumn{2}{c|}{Acc} & IoU   & Acc   & mIoU  & mAcc  & aAcc \\
\hline
    WD(2-classes) & \multicolumn{2}{c}{99.45} & \multicolumn{2}{c|}{99.60} & 79.59 & 89.15 & 89.52 & 94.38 & 98.11 \\
    \hline
    \end{tabular}%
    }
  \label{tab:wd_res}%
\end{table}%
\begin{figure}[!b]
  \centering
  \begin{minipage}[b]{0.49\linewidth}
    \includegraphics[width=1\linewidth]{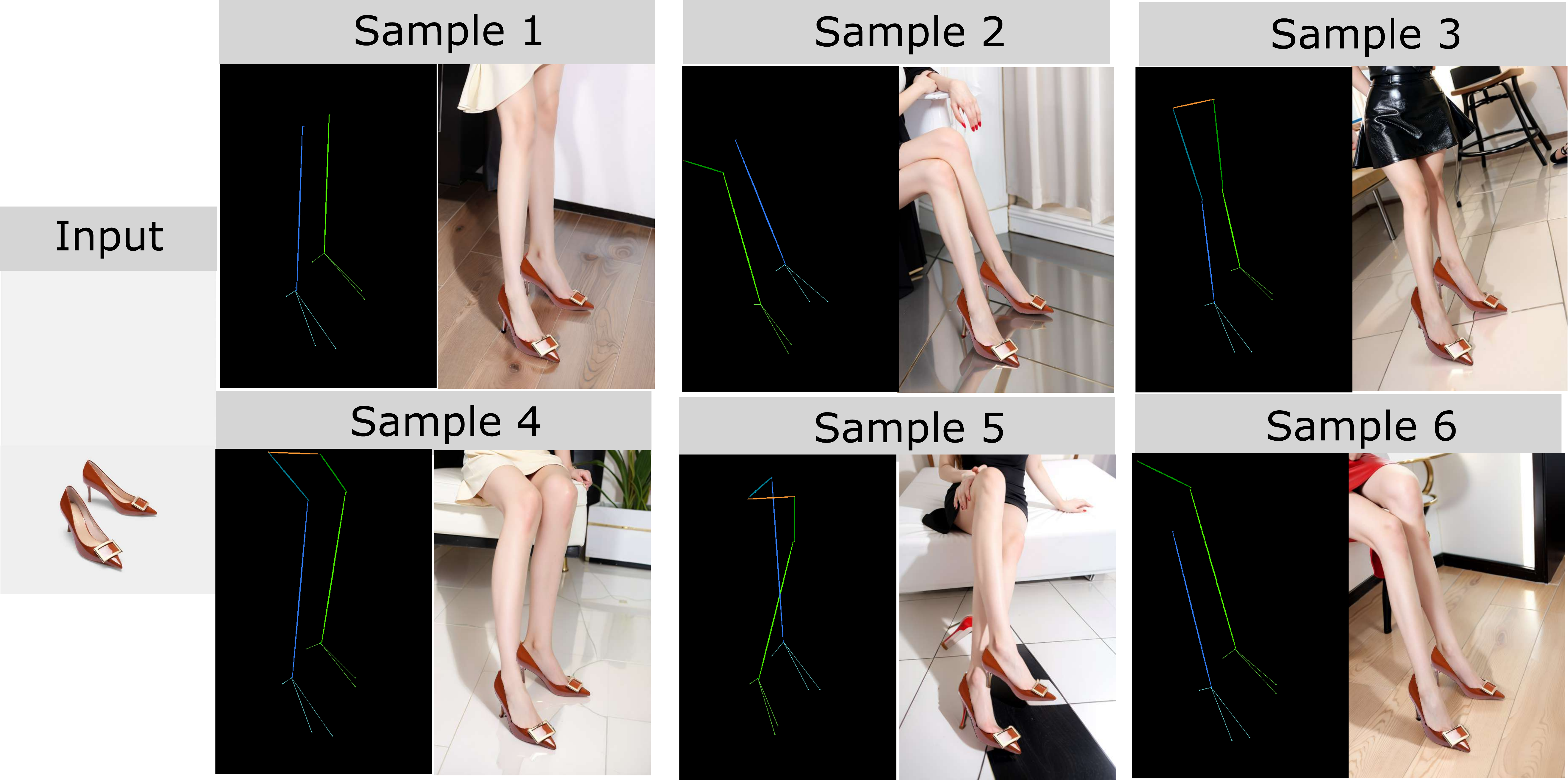}
  \caption{Diverse poses predicted by LpS and the corresponding final shoe-wearing images.}\label{fig_diversepose}
  \end{minipage}
  \begin{minipage}[b]{0.49\linewidth}
    \includegraphics[width=1\linewidth]{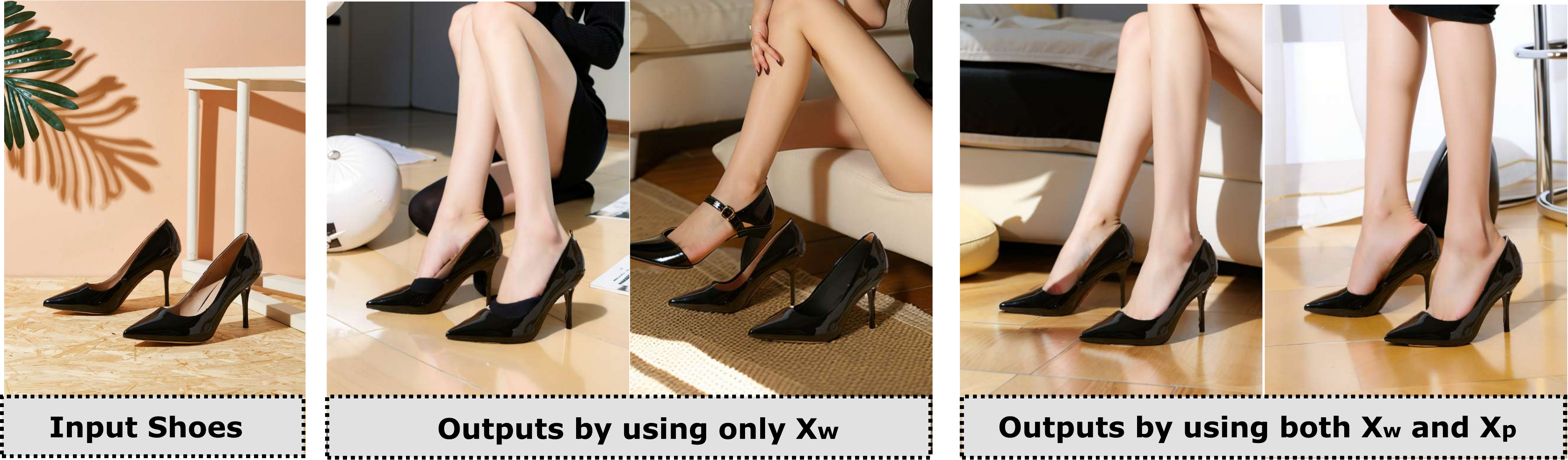}
  \caption{\small{Ablation study on $X_{p}$, one can observe that without using the predicted $X_{p}$ the generated legs will be with unreasonable poses and the interaction between human and shoes will be implausible.}}
  \label{fig_1}
  \end{minipage}
\end{figure}
\textbf{Diversity of pose synthesis:} To show the effectiveness of the LpS module, we provide various pose synthesis results based on the same one input. Figure \ref{fig_diversepose} shows that by using LpS module, the generated poses can be diverse and reasonable. And thanks to our LpS module, the final shoe-wearing images are diverse.

\begin{figure}[t!]
  \centering
  \includegraphics[width=1\linewidth]{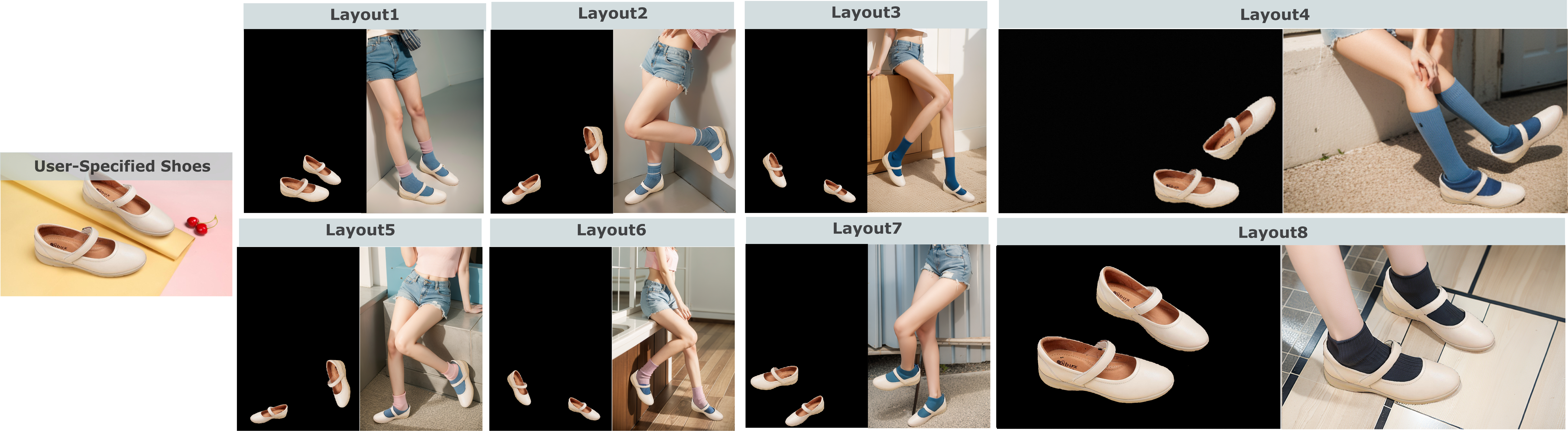}
  \caption{Results of the customized shoes layout. Best viewed with zoom-in.}\label{fig_layout}
\end{figure}

\textbf{Customized shoes layout:} While the LpS module is capable of generating diverse pose results, the range of predicted poses may be constrained due to the fixed layout of the given shoes. To enhance the diversity of outcomes, we can artificially modify the input layout of the user-specified shoes. Figure \ref{fig_layout} illustrates this concept, where we alter the layout of the given shoes to obtain a wider range of diverse results. These examples highlight the flexibility of our approach for real-life applications.

\textbf{Importance of $\boldsymbol{X}_{p}$:} In order to verify the importance of LpS module, we train another SW module by only accepting $X_{w}$ as input condition. The visual comparison results are in Fig \ref{fig_1}. From this figure, one can observe that if only $X_{w}$ is given the generated legs will be with unreasonable poses and the shoes might be not worn. While our LpS module can lead the SW module to generate images with diverse and reasonable human-leg poses. These findings align with the outcomes observed in comparisons involving painting-based methods in Section \ref{sec_qualitative}, further demonstrating the superiority and importance of our ShoeModel system.

\section{Discussion}
\label{sec:conclusion}
\textbf{Conclusion: }In this paper, we propose a novel \emph{ShoeModel} system for generating hyper-realistic advertising images of shoes, in which shoes are interacted with human-legs. To achieve this, three modules are introduced, such as wearable-area detection module, leg-pose synthesis module and shoe-wearing image generation module. Moreover, we also build the corresponding dataset, \ie, shoe-wearing dataset, to support the training of our \emph{ShoeModel}. To our best knowledge, it is the first work to learn to wear on the user-specified shoes by diffusion model. And all the experiments demonstrate the effectiveness and superiority of the proposed method. 

\section*{Acknowledgement}This work was supported in part by the National Key R$\&$D Program of China (2021YFF0900500), and the National Natural Science Foundation of China (NSFC) under grants 62441202.

%


%
%
\bibliographystyle{splncs04}
\bibliography{main}

\begin{thebibliography}{10}
\providecommand{\url}[1]{\texttt{#1}}
\providecommand{\urlprefix}{URL }
\providecommand{\doi}[1]{https://doi.org/#1}

\bibitem{binkowski2018demystifying}
Bi{\'n}kowski, M., Sutherland, D.J., Arbel, M., Gretton, A.: Demystifying mmd
  gans. arXiv preprint arXiv:1801.01401  (2018)

\bibitem{bochkovskiy2020yolov4}
Bochkovskiy, A., Wang, C.Y., Liao, H.Y.M.: Yolov4: Optimal speed and accuracy
  of object detection. arXiv preprint arXiv:2004.10934  (2020)

\bibitem{chen2024virtualmodel}
Chen, B., Zhong, C., Xiang, W., Geng, Y., Xie, X.: Virtualmodel: Generating
  object-id-retentive human-object interaction image by diffusion model for
  e-commerce marketing  (2024)

\bibitem{ge2021yolox}
Ge, Z., Liu, S., Wang, F., Li, Z., Sun, J.: Yolox: Exceeding yolo series in
  2021. arXiv preprint arXiv:2107.08430  (2021)

\bibitem{heusel2017gans}
Heusel, M., Ramsauer, H., Unterthiner, T., Nessler, B., Hochreiter, S.: Gans
  trained by a two time-scale update rule converge to a local nash equilibrium.
  Advances in neural information processing systems  \textbf{30} (2017)

\bibitem{ho2020denoising}
Ho, J., Jain, A., Abbeel, P.: Denoising diffusion probabilistic models.
  Advances in neural information processing systems  \textbf{33},  6840--6851
  (2020)

\bibitem{ho2021classifier}
Ho, J., Salimans, T.: Classifier-free diffusion guidance. In: NeurIPS 2021
  Workshop on Deep Generative Models and Downstream Applications (2021)

\bibitem{jin2020whole}
Jin, S., Xu, L., Xu, J., Wang, C., Liu, W., Qian, C., Ouyang, W., Luo, P.:
  Whole-body human pose estimation in the wild. In: Computer Vision--ECCV 2020:
  16th European Conference, Glasgow, UK, August 23--28, 2020, Proceedings, Part
  IX 16. pp. 196--214. Springer (2020)

\bibitem{ju2023humansd}
Ju, X., Zeng, A., Zhao, C., Wang, J., Zhang, L., Xu, Q.: Humansd: A native
  skeleton-guided diffusion model for human image generation. arXiv preprint
  arXiv:2304.04269  (2023)

\bibitem{kirillov2023segment}
Kirillov, A., Mintun, E., Ravi, N., Mao, H., Rolland, C., Gustafson, L., Xiao,
  T., Whitehead, S., Berg, A.C., Lo, W.Y., et~al.: Segment anything. arXiv
  preprint arXiv:2304.02643  (2023)

\bibitem{li2023gligen}
Li, Y., Liu, H., Wu, Q., Mu, F., Yang, J., Gao, J., Li, C., Lee, Y.J.: Gligen:
  Open-set grounded text-to-image generation. In: Proceedings of the IEEE/CVF
  Conference on Computer Vision and Pattern Recognition. pp. 22511--22521
  (2023)

\bibitem{mou2023t2i}
Mou, C., Wang, X., Xie, L., Zhang, J., Qi, Z., Shan, Y., Qie, X.: T2i-adapter:
  Learning adapters to dig out more controllable ability for text-to-image
  diffusion models. arXiv preprint arXiv:2302.08453  (2023)

\bibitem{nichol2021glide}
Nichol, A., Dhariwal, P., Ramesh, A., Shyam, P., Mishkin, P., McGrew, B.,
  Sutskever, I., Chen, M.: Glide: Towards photorealistic image generation and
  editing with text-guided diffusion models. arXiv preprint arXiv:2112.10741
  (2021)

\bibitem{podell2023sdxl}
Podell, D., English, Z., Lacey, K., Blattmann, A., Dockhorn, T., M{\"u}ller,
  J., Penna, J., Rombach, R.: Sdxl: Improving latent diffusion models for
  high-resolution image synthesis. arXiv preprint arXiv:2307.01952  (2023)

\bibitem{radford2021learning}
Radford, A., Kim, J.W., Hallacy, C., Ramesh, A., Goh, G., Agarwal, S., Sastry,
  G., Askell, A., Mishkin, P., Clark, J., et~al.: Learning transferable visual
  models from natural language supervision. In: International conference on
  machine learning. pp. 8748--8763. PMLR (2021)

\bibitem{ramesh2022hierarchical}
Ramesh, A., Dhariwal, P., Nichol, A., Chu, C., Chen, M.: Hierarchical
  text-conditional image generation with clip latents. arXiv preprint
  arXiv:2204.06125  \textbf{1}(2), ~3 (2022)

\bibitem{rombach2022high}
Rombach, R., Blattmann, A., Lorenz, D., Esser, P., Ommer, B.: High-resolution
  image synthesis with latent diffusion models. In: Proceedings of the IEEE/CVF
  conference on computer vision and pattern recognition. pp. 10684--10695
  (2022)

\bibitem{ruiz2023dreambooth}
Ruiz, N., Li, Y., Jampani, V., Pritch, Y., Rubinstein, M., Aberman, K.:
  Dreambooth: Fine tuning text-to-image diffusion models for subject-driven
  generation. In: Proceedings of the IEEE/CVF Conference on Computer Vision and
  Pattern Recognition. pp. 22500--22510 (2023)

\bibitem{song2020denoising}
Song, J., Meng, C., Ermon, S.: Denoising diffusion implicit models. arXiv
  preprint arXiv:2010.02502  (2020)

\bibitem{xiao2018simple}
Xiao, B., Wu, H., Wei, Y.: Simple baselines for human pose estimation and
  tracking. In: Proceedings of the European conference on computer vision
  (ECCV). pp. 466--481 (2018)

\bibitem{xie2021segformer}
Xie, E., Wang, W., Yu, Z., Anandkumar, A., Alvarez, J.M., Luo, P.: Segformer:
  Simple and efficient design for semantic segmentation with transformers.
  Advances in Neural Information Processing Systems  \textbf{34},  12077--12090
  (2021)

\bibitem{xu2022vitpose}
Xu, Y., Zhang, J., Zhang, Q., Tao, D.: Vitpose: Simple vision transformer
  baselines for human pose estimation. Advances in Neural Information
  Processing Systems  \textbf{35},  38571--38584 (2022)

\bibitem{xue2024strictly}
Xue, Y., Chen, B., Geng, Y., Xie, X., Chen, J., Ma, H.: Strictly-id-preserved
  and controllable accessory advertising image generation  (2024)

\bibitem{yang2023paint}
Yang, B., Gu, S., Zhang, B., Zhang, T., Chen, X., Sun, X., Chen, D., Wen, F.:
  Paint by example: Exemplar-based image editing with diffusion models. In:
  Proceedings of the IEEE/CVF Conference on Computer Vision and Pattern
  Recognition. pp. 18381--18391 (2023)

\bibitem{ye2023ip-adapter}
Ye, H., Zhang, J., Liu, S., Han, X., Yang, W.: Ip-adapter: Text compatible
  image prompt adapter for text-to-image diffusion models  (2023)

\bibitem{zhang2023adding}
Zhang, L., Rao, A., Agrawala, M.: Adding conditional control to text-to-image
  diffusion models. In: Proceedings of the IEEE/CVF International Conference on
  Computer Vision. pp. 3836--3847 (2023)

\bibitem{zhang2023controlvideo}
Zhang, Y., Wei, Y., Jiang, D., Zhang, X., Zuo, W., Tian, Q.: Controlvideo:
  Training-free controllable text-to-video generation. arXiv preprint
  arXiv:2305.13077  (2023)

\bibitem{zhu2023h3wb}
Zhu, Y., Samet, N., Picard, D.: H3wb: Human3. 6m 3d wholebody dataset and
  benchmark. In: Proceedings of the IEEE/CVF International Conference on
  Computer Vision. pp. 20166--20177 (2023)

\end{thebibliography}
\end{document}